%% file: GMM_RNTR_arxiv.tex
\documentclass[12pt,a4paper]{article}

\usepackage{amsfonts,amsmath,amssymb} 
\usepackage[pdftex]{graphicx} 
\usepackage[utf8]{inputenc}
\usepackage{verbatim}
\usepackage{hyperref}
\hypersetup{
	colorlinks=true, 
	linkcolor=black,  
	citecolor=black,
	urlcolor=black
}
\newtheorem{Theorem}{Theorem}
\usepackage{stmaryrd}
\usepackage[linesnumbered,ruled]{algorithm2e}
\usepackage{subcaption}

\usepackage{tikz}
\usetikzlibrary{positioning,chains,fit,shapes,calc}
\usepackage{multicol}
\usepackage{booktabs}

\usepackage{placeins}
\usepackage{mathtools}

\usepackage[comma,authoryear,round]{natbib}

\usepackage{chngcntr}
\counterwithin{table}{section}
\counterwithin{figure}{section}

\newcommand{\R}{\mathbb{R}} 
\newcommand{\N}{\mathbb{N}} 
\newcommand{\bP}{\mathbb{P}} 
\newcommand{\bS}{\mathbb{S}} 
\newcommand{\Deriv}{\text{D}}
\newcommand{\tr}{\text{tr}}

\newcommand{\Pen}{\text{ Pen}}

\newcommand{\norm}[1]{\left\lVert#1\right\rVert}
\newcommand{\Mf}{{\mathcal{M}}}       							
\def\one{\mbox{1\hspace{-4.25pt}\fontsize{12}{14.4}\selectfont\textrm{1}}}

\DeclareMathOperator{\diag}{diag}
\DeclareMathOperator{\grad}{grad}
\DeclareMathOperator{\Hess}{Hess}

\newcommand*{\qed}{\null\nobreak\hfill\ensuremath{\square}}

\RequirePackage{fix-cm}
\providecommand{\keywords}[1]
{
	\small	
	\textbf{\textit{Keywords---}} #1
}

\begin{document}

\title{A Riemannian Newton Trust-Region Method for Fitting Gaussian Mixture Models}
 
 \author{
        Lena Sembach\thanks{University of Trier,
        	FB IV - Department of Mathematics, D-54296, Trier,
        	Germany ({\tt sembach@uni-trier.de}).} 
         \and
         Jan Pablo Burgard\thanks{University of Trier,
        FB IV - Department of Economics, D-54296, Trier,
        Germany ({\tt burgardj@uni-trier.de}).}
        \and        
        Volker Schulz\thanks{University of Trier,
        FB IV - Department of Mathematics, D-54296, Trier,
        Germany ({\tt volker.schulz@uni-trier.de}).}
        }

\date{}

\maketitle

\begin{abstract}
	Gaussian Mixture Models are a powerful tool in Data Science and Statistics that are mainly used for clustering and density approximation. The task of estimating the model parameters is in practice often solved by the Expectation Maximization (EM) algorithm which has its benefits in its simplicity and low per-iteration costs. However, the EM converges slowly if there is a large share of hidden information or overlapping clusters. Recent advances in manifold optimization for Gaussian Mixture Models have gained increasing interest. We introduce an explicit formula for the Riemannian Hessian for Gaussian Mixture Models. On top, we propose a new Riemannian Newton Trust-Region method which outperforms current approaches both in terms of runtime and number of iterations. 
	We apply our method on clustering problems and density approximation tasks. Our method is very powerful for data with a large share of hidden information compared to existing methods.
\end{abstract}

\section{Introduction}
\label{intro}
\input{./Chapters/Introduction}

\section{Riemannian Setting for Gaussian Mixture Models}
\label{sec: 2}
\input{./Chapters/RiemSetting_GMM}

\section{Riemannian Newton Trust-Region Algorithm}
\label{sec: 3}
\input{./Chapters/RNTR}

\section{Numerical Results}
\label{sec: 4}
\input{./Chapters/NumericalResults}

\newpage
\section{Conclusion}
\label{sec: Conclusion}
\input{./Chapters/Conclusion}

\FloatBarrier

\section*{Acknowledgments}
This research has been supported by the German Research Foundation (DFG) within the Research Training Group 2126:  Algorithmic Optimization, Department of Mathematics, University of Trier, Germany. We thank Mrs. Ngan Mach for her support for numerical results.

%
%


\bibliographystyle{abbrvnat}
\bibliography{ref}

%
%

\newpage
\appendix
\section{Unboundedness of the Reformulated Problem}
\label{app: unbounded}
\input{./Chapters/Appendix/Unboundedness}

\newpage

\section{Truncated Conjugate Gradient Method}
\label{app:tCG}
\input{./Chapters/Appendix/tCG}

\newpage
\section{Additional Numerical Results}
\label{app:data}
We present numerical results additional to Section \ref{sec: 4} both for simulated clustering data and for real-world data.

\subsection{Simulated data}
\input{./Chapters/Appendix/AdditionalNumerical}

\newpage
\subsection{Real-world datasets}
\input{./Chapters/Appendix/Realworld}



\end{document}

%% file: Chapters/Introduction.tex
Gaussian Mixture Models are widely recognized in Data Science and Statistics. 
	The fact that any probability density can be approximated by a Gaussian Mixture Model with a sufficient number of components makes it an attractive tool in statistics. However, this comes with some computational limitations, where some of them are described in \cite{Ormoneit_Tresp, Lee_McLachlan_2013, Coretto}. Nevertheless, we here focus on the benefits of Gaussian mixture models. Besides the goal of density approximation, the possibility of modeling latent features by the underlying components make it also a strong tool for soft clustering tasks.
Typical applications are to be found in the area of image analysis \citep{Applic_ImageSeg,Applic_ImageAnom, Zoran}, pattern recognition \citep{Applic_patternrec,bishop:2006:PRML}, econometrics \citep{Applic_Rents,Compiani} and many others.

We state the Gaussian Mixture Model in the following: \\
Let $K \in \N$ be given. The Gaussian Mixture Model (with $K$ components) is given by the (multivariate) density function $p$ of the form
\begin{align}
p(x) = \sum\limits_{j=1}^K \alpha_j p_{\mathcal{N}}(x; \mu_j, \Sigma_j), 	\qquad x \in \R^d,
\label{prob_dens}
\end{align}
with positive mixture components $\alpha_j$ that sum up to $1$ and Gaussian density functions $p_{\mathcal{N}}$ with means $\mu_j \in \R^d$ and covariance matrices $\Sigma_j \in \R^{d \times d}$. In order to have a well-defined expression, we impose $\Sigma_j \succ 0$, i.e. the $\Sigma_j$ are symmetric positive definite.	

Given observations $x_1, \dots, x_m$, the goal of parameter estimation for Gaussian Mixture Models consists in maximizing the log-likelihood. This yields the optimization problem

\begin{align}
\max_{\substack{{\alpha} \in \Delta_K \\ \mu_j \in \R^d \\ \Sigma_j \succ 0}} \sum\limits_{i=1}^m \log \left( \sum\limits_{j=1}^K \alpha_j p_{\mathcal{N}}(x_i; \mu_j, \Sigma_j)\right),			
\label{opt_original}
\end{align}
where 
\begin{align*}
\Delta_K = \{(\alpha_1,\dots, \alpha_K), \text{ } \alpha_j \in \R^{+} \forall j, \quad  \sum\limits_{j=1}^K \alpha_j =1\} 
\end{align*} 
is the K-dimensional probability simplex and the covariance matrices $\Sigma_j$ are restricted to the set of positive definite matrices. \\

In practice, this problem is commonly solved by the Expectation Maximization (EM) algorithm. It is known that the Expectation Maximization algorithm converges fast if the $K$ clusters are well separated. \cite{Ma} showed that in such a case, the convergence rate is superlinear. However, Expectation Maximization has its speed limits for highly overlapping clusters, where the latent variables have a non-neglibible large probability among more than one cluster. In such a case, the convergence is linear \citep{Xu} which might results in slow parameter estimation despite very low per-iteration costs. 

From a nonlinear optimization perspective, the problem in \eqref{opt_original} can be seen as a constrained nonlinear optimization problem. However, the positive definiteness constraint of the covariance matrices $\Sigma_j$ is a challenge for applying standard nonlinear optimization algorithms. While this constraint is naturally fulfilled in the EM algorithm, we cannot simply drop it as we might leave the parameter space in alternative methods. Approaches to this problem like introducing a Cholesky decomposition \citep{Salakhutdinov_CG,Xu}, or using interior point methods via smooth convex inequalities \citep{Vanderbei} can be applied and one might hope for faster convergence with Newton-type algorithms. Methods like using a Conjugate Gradient Algorithm (or a combination of both EM and CG) led to faster convergence for highly overlapping clusters \citep{Salakhutdinov_CG}. However, this induces an additional numeric overhead by imposing the positive-definiteness constraint by a Cholesky decomposition. 

In recent approaches \cite{Hosseini15,Hosseini20} suggest to exploit the geometric structure of the set of positive definite matrices. As an open set in the set of symmetric matrices, it admits a manifold structure \citep{Bhatia} and thus the concepts of Riemannian optimization can be applied. The concept of Riemannian optimization, i.e. optimizing over parameters that live on a smooth manifold is well studied and has gained increasing interest in the domain of Data Science, for example for tensor completion problems \citep{Heidel}. However, the idea is quite new for Gaussian Mixture Models and \cite{Hosseini15,Hosseini20} showed promising results with a Riemannian LBFGS and Riemannian Stochastic Gradient Descent Algorithm. The results in \cite{Hosseini15,Hosseini20} are based on a reformulation of the log-likelihood in \eqref{opt_original} that turns out to be very efficient in terms of runtime. By design, the algorithms investigated in \cite{Hosseini15,Hosseini20} do not use exact second-order information of the objective function.
Driven by the quadratic local convergence of the Riemannian Newton method, we thus might hope for faster algorithms with the availability of the Riemannian Hessian. In the present work, we derive a formula for the Riemannian Hessian of the reformulated log-likelihood and suggest a Riemannian Newton Trust-Region method for parameter estimation of Gaussian Mixture Models.\\

The paper is organized as follows. In Section \ref{sec: 2}, we introduce the reader to the concepts of Riemannian optimization and the Riemannian setting of the reformulated log-likelihood for Gaussian Mixture Models. In particular, we derive the expression for the Riemannian Hessian in Subsection \ref{subsec:RiemSettGMM} which is a big contribution to richer Riemannian methods for Gaussian Mixture Models. In Section \ref{sec: 3} we present the Riemannian Newton Trust-Region Method and prove the global convergence and superlinear local convergence for our problem. We compare our proposed method with existing algorithms both on artificial and real world data sets in Section \ref{sec: 4} for the task of clustering and density approximation.

%% file: Chapters/RiemSetting_GMM.tex
We will build the foundations of Riemannian Optimization in the following to specify the characteristics for Gaussian Mixture Models afterwards. In particular, we introduce a formula for the Riemannian Hessian for the reformulated problem which is the basis for second-order optimization algorithms.

\subsection{Riemannian Optimization}
\label{subsec:RO}
To construct Riemannian Optimization methods, we briefly state the main concepts of Optimization on Manifolds or Riemannian Optimization. A good introduction is \cite{Absil} and \cite{boumal2020intromanifolds}, we here follow the notations of \cite{Absil}. The concepts of Riemannian Optimization are based on concepts from unconstrained Euclidean optimization algorithms and are generalized to (possibly nonlinear) manifolds.\\

\begin{figure}\begin{center}
		\begin{tikzpicture}[scale=0.25]
		
		\filldraw (0,0) circle (1.35pt);
		\filldraw (14.01,-1.485) circle (1.05pt);
		\filldraw (8.195,-2.585) circle (1.15pt);
		
		\draw [line width=0.5mm, black ] (0,0) to [bend left=45] (7.8,6.3)
		to [bend left=45] (14,-1.5);
		\draw [line width=0.5mm, black ] (1.035,3) to [bend left=55] (7.1,1.7)
		to [bend left=12] (8.2,-2.6);
		\draw [line width=0.5mm, black ] (8.2,-2.57) to [bend left=25] (14,-1.47);
		\draw [line width=0.5mm, black ] (0,0) to [bend left=12] (6.1,3);
		\node at (6.5,-1) {$\Mf$};
		\filldraw (7,5) circle (4pt);				
		\node at (6,5.5) {$\theta$};

		\draw [line width=0.3mm] (4,4) -- ++(10,-2.5);
		\draw [line width=0.3mm] (4,4) -- ++(2.5,4);
		\draw [line width=0.3mm] (6.5,8) -- ++(10,-2.5);
		\draw [line width=0.3mm] (14,1.5) -- ++(2.5,4);
		\filldraw (4,4) circle (0.4pt); \filldraw (14,1.5) circle (0.4pt);
		\filldraw (6.5,8) circle (0.4pt); \filldraw (16.5,5.5) circle (0.4pt);
		\node at (18,4) {$T_{\theta} \Mf$};

		\draw [->,line width=0.5mm](7,5) -- ++(4,-1);	
		\node at (11.5,3.5) {$\xi_{\theta}$};

		\draw [dotted,line width=0.5mm] (7,5) to [bend left=25] node{.}(11,1);
		\filldraw (11,1) circle (4pt);
		\node at (11,0) {$R_{\theta}(\xi_{\theta})$};
		
		\end{tikzpicture}\caption{Retraction-based Riemannian Optimization}\label{Fig:retr_based_opt}
	\end{center}
\end{figure}
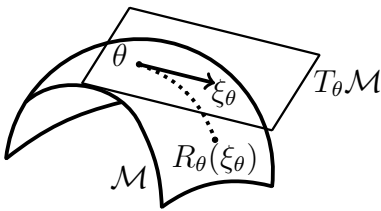

	A manifold is a space that locally resembles Euclidean space, meaning that we can locally map points on manifolds to $\R^n$ via bicontinuous mappings. Here, $n$ denotes the dimension of the manifold. In order to define a generalization of differentials, Riemannian optimization methods require smooth manifolds meaning that the transition mappings are smooth functions. As manifolds are in general not vector spaces, standard optimization algorithms like line-search methods cannot be directly applied as the iterates might leave the admissible set. Instead, one moves along tangent vectors in tangent spaces $T_{\theta}\Mf$, local approximations of a point $\theta$ on the manifold, i.e. $\theta \in \Mf$. Tangent spaces are basically first-order approximations of the manifold at specific points and the \textit{tangent bundle} $T\Mf$ is the disjoint union of the tangent spaces $T_{\theta}\Mf$. In Riemannian manifolds, each of the tangent spaces $T_{\theta}\Mf$ for $\theta \in \Mf$ is endowed with an inner product $\langle  \cdot, \cdot \rangle_{\theta}$ that varies smoothly with $\theta$. The inner product is essential for Riemannian optimization methods as it admits some notion of length associated with the manifold. The optimization methods also require some local pull-back from the tangent spaces $T_{\theta} \Mf$ to the manifold $\Mf$ which can be interpreted as moving along a specific curve on $\Mf$ (dotted curve in Figure \ref{Fig:retr_based_opt}). This is realized by the concept of \textit{retractions}: Retractions are mappings from the tangent bundle $T\Mf$ to the manifold $\Mf$ with rigidity conditions: we move through the zero element $0_{\theta}$ with velocity $\xi_{\theta} \in T_{\theta}\Mf$ , i.e. $DR_{\theta} ( 0_{\theta} )[ \xi_{\theta} ] = \xi_{\theta}.$ Furthermore, the retraction of $0_{\theta} \in T_{\theta}\Mf$ at $\theta$ is $\theta$ itself (see Figure \ref{Fig:retr_based_opt}).

\newpage
Roughly spoken, a step of a Riemannian optimization algorithm works as follows:
\begin{itemize}
	\item At iterate $\theta^t$, take a new step $\xi_{\theta^t}$ on the tangent space $T_{\theta^t}\Mf$
	\item Pull back the new step to the manifold by applying the retraction at point $\theta^t$ by setting $\theta^{t+1} = R_{\theta^t} (\xi_{\theta^t})$
\end{itemize}

Here, the crucial part that has an impact on convergence speed is updating the new iterate on the tangent space, just like in the Euclidean case. As Riemannian optimization algorithms are a generalization of Euclidean unconstrained optimization algorithms, we thus introduce a generalization of the gradient and the Hessian.

\paragraph{Riemannian Gradient.}
In order to characterize Riemannian gradients, we need a notion of differential of functions defined on manifolds.\\

The \textit{differential} of $f : \Mf \rightarrow \R$  at $\theta$ is the linear operator
$\Deriv f(\theta) : T\Mf_{\theta} \rightarrow \R$ defined by:
\begin{align*}
\Deriv f (\theta)[v] = \frac{d}{dt} f(c(t))\biggr\rvert_{t=0},	
\end{align*}
where $c: I \rightarrow \Mf$, $0 \in I \subset \R$ is a smooth curve on $\Mf$ with $c'(0) = v$.\\

The Riemannian gradient can be uniquely characterized by the differential of the function $f$ and the inner product associated with the manifold:\\

The \textit{Riemannian gradient} of a smooth function $f: M \rightarrow \R$ on a Riemannian manifold is a mapping $\grad f: \Mf \rightarrow T\Mf$ such that, for all $\theta \in \Mf$, $\grad f(\theta)$ is the unique tangent vector in $T_{\theta}\Mf$ satisfying
\begin{align*}
\langle \grad f(\theta), \xi_{\theta} \rangle_{\theta} = \Deriv f(\theta) [\xi_{\theta}] 	\quad \forall \xi_{\theta} \in \Mf.
\end{align*}

\paragraph{Riemannian Hessian.} We can also generalize the Hessian to its Riemannian version. To do this, we need a tool to differentiate along tangent spaces, namely the Riemannian connection (for details see \cite[Section 5.3]{Absil}).\\

The \textit{Riemannian Hessian} of $f: \Mf \rightarrow \R$ at $\theta$ is the linear operator $\Hess f(\theta): T_{\theta}\Mf \rightarrow T_{\theta}\Mf$ defined by
\begin{align*}
\Hess f(\theta)[\xi_{\theta}] = \nabla_{\xi_{\theta}} \grad f(\theta),
\end{align*}
where $\nabla$ is the Riemannian connection with respect to the Riemannian manifold.

\subsection{Reformulation of the Log-likelihood}
\label{subsec:Refo}

In \cite{Hosseini15}, the authors experimentally showed that applying the concepts of Riemannian optimization to the objective in \eqref{opt_original} cannot compete with Expectation Maximization. This can be mainly led back to the fact that the maximization in the M-step of EM, i.e. the maximization of the log-likelihood for a single Gaussian, is a concave problem and easy to solve, it even admits a closed-form solution. Nevertheless, when considering Riemannian optimization for \eqref{opt_original}, the maximization of the log-likelihood of a single Gaussian is not geodesically concave (concavity along the shortest curve connecting two points on a manifold). The following reformulation introduced by \cite{Hosseini15} removes this geometric mismatch and results in a speed-up for the Riemannian algorithms.\\

We augment the observations $x_i$ by introducing the observations $y_i = (x_i, 1)^T \in \R^{d+1}$ for $i=1, \dots, m$ and consider the optimization problem 
	\begin{align}
	\max_{\theta = ((S_1,\dots, S_K), (\eta_1,\dots, \eta_{K-1}))} \hat{\mathcal{L}} (\theta) = \sum\limits_{i=1}^m \log \left(\sum\limits_{j=1}^K h^i(\theta_j)\right), 
	\label{opt_ref}
	\end{align}
where
	\begin{align}
	h^i(\theta_j) &= \frac{\exp(\eta_j)}{\sum\limits_{k=1}^K \exp(\eta_k)} \frac{ \exp \left(\frac{1}{2}\left(1-y_i^T S_j^{-1} y_i \right)\right)}{\sqrt{(2\pi)^d \det(S_j)}} =  \frac{\exp(\eta_j)}{\sum\limits_{k=1}^K \exp(\eta_j)}q_{\mathcal{N}}(y_i;S_j),
	\label{h}
	\end{align}	
	with $q_{\mathcal{N}}(y_i;S_j) = \sqrt{2\pi} \exp(\frac{1}{2}) p_{\mathcal{N}}(y_i;0,S_j)$  for parameters ${\theta_j = (S_j, \eta_j)}$, $j=1, \dots, K-1$ and ${\theta_K = (S_K, 0)}$.

This means that instead of considering Gaussians of $d$ -dimensional variables $x$, we now consider Gaussians of $d+1$ -dimensional variables $y$ with zero mean and covariance matrices $S_j$. The reformulation leads to faster Riemannian algorithms, as it has been shown in \cite{Hosseini15} that the maximization of a single Gaussian
\begin{align*}
\max_{S \succ 0} \sum\limits_{i=1}^m \log q_{\mathcal{N}}(y_i; S)
\end{align*}
is geodesically concave.\\

Furthermore, this reformulation is faithful, as the original problem \eqref{opt_original} and the reformulated problem \eqref{opt_ref} are equivalent in the followings sense:

\begin{Theorem}{\citep[Theorem 2.2]{Hosseini15}}\label{Th: equiv_opt} 
	A local maximum of the reformulated GMM log-likelihood $ \hat{\mathcal{L}} (\theta)$ with maximizer $\theta^{*} = ({\theta}_1^{*},\dots,{\theta}_K^{*})$, ${\theta}_j^{*} = ({S}_j^{*}, {\eta}_j^{*})$ is a local maximum of the original log-likelihood $\mathcal{L}((\alpha_j, \mu_j,\Sigma_j)_{j})$ with maximizer $({\alpha}_j^{*}, {\mu}_j^{*}, {\Sigma}_j^{*})_{j}$. Here, $\mathcal{L}$ denotes the objective in the problem \eqref{opt_original}.
\end{Theorem}
The relationship of the maximizers is given by
\begin{align}
S_j^{*} &= \left( \begin{array}{cc}
\Sigma_j^{*} + \mu_j^{*}{\mu_j^{*}}^T & \quad \mu_j^{*} \label{ref_1}\\
{\mu_j^{*}}^T & \quad 1\\
\end {array} \right),\\
\eta_j^{*} &= \log\left(\frac{\alpha_j^{*}}{\alpha_K^{*}}\right) \quad j=1, \dots, K-1; \quad \eta_K \equiv 0 \label{ref_2}.
\end{align}
This means that instead of solving the original optimization problem \eqref{opt_original} we can easily solve the reformulated problem \eqref{opt_ref} on its according parameter space and transform the optima back by the relationships \eqref{ref_1} and \eqref{ref_2}.

\paragraph{Penalizing the objective.}
When applying Riemannian optimization algorithms on the reformulated problem \eqref{opt_ref}, covariance singularity is a challenge. Although this is not observed in many cases in practice, it might result in unstable algorithms. This is due to the fact that the objective in \eqref{opt_ref} is unbounded from above, see Appendix \ref{app: unbounded} for details. The same problem is extensively studied for the original problem \eqref{opt_original} and makes convergence theory hard to investigate. An alternative consists in considering the maximum a posterior log-likelihood for the objective. If conjugate priors are used for the variables $\mu_j, \Sigma_j$, the optimization problem remains structurally unchanged and results in a bounded objective (see \cite{SnoussiDjaf}). Adapted versions of Expectation Maximization have been proposed in the literature and are often applied in practice.\\

A similar approach has been proposed in \cite{Hosseini20} where the reformulated objective \eqref{opt_ref} is penalized by an additive term that consists of the logarithm of the Wishart prior, i.e. we penalize each of the $K$ components with
\begin{align}
\psi(S_j, \Psi) = -\frac{\rho}{2} \log\det(S_j) - \frac{\beta}{2} \tr(\Psi {S_j}^{-1}),
\label{pen_mat}
\end{align}
where $\Psi$ is the block matrix
\begin{align*}
\Psi =  \left(\begin{array}{cc} \frac{\gamma}{\beta}\Lambda + \kappa \lambda \lambda^T & \kappa \lambda \\
\kappa \lambda^T & \kappa \end{array}\right)
\end{align*}
for $\lambda \in \R^d$, $\Lambda \in \R^{d \times d}$ and $\gamma, \beta, \rho, \nu, \kappa \in \R$. If we assume that ${\rho = \gamma(d+\nu + 1) + \beta}$, the results of Theorem \ref{Th: equiv_opt} are still valid for the penalized version, see \cite{Hosseini20}. Besides, the authors introduce an additive term to penalize very tiny clusters by introducing Dirichlet priors for the mixing coefficients $\alpha_j = \frac{\exp(\eta_j)}{\sum\limits_{k=1}^K \exp(\eta_k)}$, i.e.
\begin{align}
\varphi(\eta, \zeta) = \zeta \left(\sum\limits_{j=1}^K \eta_j - K\log\left(\sum\limits_{k=1}^K \exp(\eta_k)\right) \right).
\label{pen_weight}
\end{align}

In total, the penalized problem is given by

\begin{align}
\max_{\theta} \hat{\mathcal{L}}_{\text{pen}} (\theta; \Psi, \zeta)  = \sum\limits_{i=1}^m \log \left(\sum\limits_{j=1}^K h^i(\theta_j)\right)+ \Pen(\theta),  
\label{problem_w_pen}
\end{align}
where
\begin{align*}
\Pen(\theta) =\sum\limits_{j=1}^K \psi(S_j, \Psi) + \varphi(\eta, \zeta).
\end{align*}

The use of such an additive penalizer leads to a bounded objective:

\begin{Theorem}
	\label{Prop: Boundedness}
	The penalized optimization problem in \eqref{problem_w_pen} is bounded from above.
\end{Theorem}

\noindent \textit{Proof.} We follow the proof for the original objective \eqref{opt_original} from \cite{SnoussiDjaf}. The penalized objective reads
\begin{align*}
\hat{\mathcal{L}}_{\text{pen}} (\theta; \Psi, \zeta) = \sum\limits_{i=1}^m \log \left(\sum\limits_{j=1}^K q_{\mathcal{N}}^{\text{AP}} (\theta_j, S_j; \Psi, \zeta) \right),
\end{align*}
where
\begin{align*}
&q_{\mathcal{N}}^{\text{AP}} (\theta_j, S_j; \Psi, \zeta) = h^i(\theta_j) \bigg(\prod_{k=1}^K  \det(S_k)^{-\frac{\rho}{2}}\exp \left(-\frac{1}{2}\tr(S_k^{-1}\Psi)\right) {\alpha_k}^{\zeta} \bigg)^{1/m}.
\end{align*}
We get the upper bound
\begin{align}
q_{\mathcal{N}}^{\text{AP}} &(\theta_j, S_j; \Psi, \zeta)  \leq h^i(\theta_j) \prod_{k=1}^K \det(S_k)^{-\frac{\rho}{2}} \exp\left(-\frac{1}{2} \tr(S_k^{-1}\Psi)\right) \alpha_k^{\zeta} \notag \\
& \leq a \alpha_j \det(S_j)^{-\frac{d}{2}} \prod_{k=1}^K  \det(S_k)^{-\frac{\rho}{2}}\exp\left(-\frac{1}{2} \tr(S_k^{-1}\Psi)\right) \notag \\
&  = a \alpha_j (\det(S_j))^{-\frac{d+\rho}{2}} \exp\left(-\frac{1}{2} \tr(S_j^{-1}\Psi)\right) \prod\limits_{\substack{k=1 \\ k \neq j}}^K  \det(S_k)^{-\frac{\rho}{2}} \exp\left(-\frac{1}{2} \tr(S_k^{-1}\Psi)\right),
\label{Pr:eq_upperbound_1}
\end{align}
where we applied Bernoulli's inequality in the first inequality and used the positive definiteness of $S_j$ in the second inequality. $a$ is a positive constant independent of $S_j$ and $S_k$.

By applying the relationship $\det(A)^{1/n} \leq \frac{1}{n} \tr(A)$ for $A \in \R^{n \times n}$ by the inequality of arithmetic and geometric means, we get for the right hand side of \eqref{Pr:eq_upperbound_1}
\begin{align}
\det(S_k)^{-\frac{b}{2}} \exp(-\frac{1}{2} \tr(S_k^{-1}\Psi)) \leq (\det(S_k))^{- \frac{b}{2}} \exp\left(-\frac{d+1}{2} \left(\frac{\det(\Psi)}{\det(S_k)}\right)^{\frac{1}{d+1}}\right)
\label{Pr:eq_upperbound_2}
\end{align}
for a constant $b > 0$. The crucial part on the right side of \eqref{Pr:eq_upperbound_2} is when one of the $S_k$ approaches a singular matrix and thus the determinant approaches zero. Then, we reach the boundary of the parameter space. We study this issue in further detail:

Without loss of generality , let $k=1$ be the component where this occurs. Let $S_1^{*}$ be a singular semipositive definite matrix of rank $r < d+1$. Then, there exists a decomposition of the form
\begin{align*}
S_1^{*} = U^T D U,
\end{align*}
where $D = \diag(0, \dots, 0, \lambda_{d-r}, \lambda_{d-r+1}, \dots, \lambda_{d+1})$, $\lambda_{l} > 0$ for $l=d-r,\dots, d+1$ and $U$ an orthogonal square matrix of size $d+1$. Now consider the sequence $S_1^{(n)}$ given by
\begin{align}
S_1^{(n)} = U^{T} D^{(n)} U, 	\label{Proof:bound_dec}
\end{align}
where 
\begin{align*}
D^{(n)} = \diag(\lambda_1^{(n)}, \dots, \lambda_{d-r-1}^{(n)},\lambda_{d-r}, \lambda_{d-r+1}, \dots, \lambda_{d+1})
\end{align*}
with $\left(\lambda_l^{(n)}\right)_{l=1, \dots, d-r-1} $ converging to $0$ as $n \rightarrow \infty$. Then, the matrix $S_1^{(n)}$ converges to $S_1^{(*)}$. \\
Setting $\lambda^{(n)} = \prod\limits_{1}^{d-r-1} \lambda_l^{(n)}$ and $\lambda^{+} = \prod\limits_{d-r}^{d+1} \lambda_l$,
the right side of \eqref{Pr:eq_upperbound_2} reads
\begin{align*}´
\left(\lambda^{(n)} \lambda^{+}\right)^{-\frac{b}{2}}\exp\left(-\frac{d+1}{2} \left(\frac{\det(\Psi)}{\lambda^{+}\lambda^{(n)}}\right)^{\frac{1}{d+1}}\right),
\end{align*}
which converges to $0$ as $n \rightarrow \infty$ by the rule of Hôpital.\qed
\\

With Theorem \ref{Prop: Boundedness}, we are able to study the convergence theory of the reformulated problem \eqref{opt_ref} in Section \ref{sec: 3}.

\subsection{Riemannian Characteristics of the reformulated Problem}
\label{subsec:RiemSettGMM}
To solve the reformulated problem \eqref{opt_ref} or the penalized reformulated problem \eqref{problem_w_pen}, we specify the Riemannian characteristics of the optimization problem. It is an optimization problem over the product manifold 
\begin{align}
\Mf = \left(\bP^{d+1}\right)^K \times \R^{K-1},
\label{manif_pr}
\end{align}
where $\bP^{d+1}$ is the set of strictly positive definite matrices of dimension $d+1$. The set of symmetric matrices is tangent to the set of positive definite matrices as $\bP^{d+1}$ is an open subset of it. Thus the tangent space of the manifold \eqref{manif_pr} is given by
\begin{align}
T_{\theta} \Mf &= \left(\bS^{d+1}\right)^K \times \R^{K-1},
\label{tang_probl}
\end{align}
where $\bS^{d+1}$ is the set of symmetric matrices of dimension $d+1$. The inner product that is commonly associated with the manifold of positive definite matrices is the \textit{intrinsic inner product} 
\begin{align}
\langle \xi_S, \chi_S \rangle_S = \tr(S^{-1}\xi_S S^{-1} \chi_S),
\label{inner_intr}
\end{align}
where $S \in \bP^{d+1}$ and $\xi_S, \chi_S \in \bS^{d+1}$. The inner product defined on the tangent space \eqref{tang_probl} is the sum over all component-wise inner products and reads
\begin{align}
\langle \xi_{\theta},\chi_{\theta} \rangle_{\theta} &=\sum\limits_{j=1}^K \tr(S_j^{-1}\xi_{S_j} S_j^{-1} \chi_{S_j}) + \xi_{\eta}^T \chi_{\eta}, 
\label{inner_probl}
\end{align}
with
\begin{align*}
\theta = ((S_1, \dots, S_K), \eta) \in \Mf, &\xi_{\theta} = \left((\xi_{S_1}, \dots, \xi_{S_K}), \xi_{\eta}\right) \in T_{\theta}\Mf,\\ &\chi_{\theta} =\left((\chi_{S_1}, \dots, \chi_{S_K}), \chi_{\eta}\right)  \in T_{\theta} \Mf.
\end{align*}
The retraction used is the exponential map on the manifold given by
\begin{align}
R_{\theta}(\xi) = \left( \begin{array}{c} \left(S_j\exp\left(S_j^{-1} \xi_ {S_j}\right)\right)_{j=1,\dots,K} \\ (\eta_j + \xi_{\eta_j})_{j=1,\dots, K-1} \end{array} \right),
\label{Retraction_probl}
\end{align}
see \cite{Jeuris}.

\paragraph{Riemannian Gradient and Hessian.} We now specify the Riemannian Gradient and the Riemannian Hessian in order to apply second-order methods on the manifold. The Riemannian Hessian in Theorem \ref{Lemma:Hess} is novel for the problem of fitting Gaussian Mixture Models and provides a way of making second-order methods applicable.
\begin{Theorem}
	The Riemannian gradient of the reformulated problem reads 
	\begin{align*}
	\grad \hat{\mathcal{L}}(\theta) = \left(
	\chi_S, \chi_{\eta} \right),
	\end{align*} where
	\begin{align*}
	\chi_S = \left(\frac{1}{2} \sum\limits_{i=1}^m f_l^i (y_i {y_i}^T - S_l)\right)_{l=1,\dots, K}, \quad \chi_{\eta} = \left(\sum\limits_{i=1}^m f_r^i - \alpha_r \sum\limits_{j=1}^K f_j^i \right) _{r=1,\dots, K-1},
	\end{align*}
	where 
	\begin{align*}
	f_l^i = \frac{h^i(\theta_l)}{\sum\limits_{j=1}^K h^i(\theta_j)}, \quad 
	\alpha_r = \frac{\exp(\eta_r)}{\sum\limits_{k=1}^K \exp(\eta_k)}.
	\end{align*}
	The additive terms for the penalizers in \eqref{pen_mat}, \eqref{pen_weight} are given by
	\begin{align*}
	\grad \Pen(\theta) = \left(\begin{array}{c} \left(-\frac{1}{2}\left(\rho S_l - \beta \Psi\right)\right)_{l=1,\dots, K} \\ \zeta \left(1- K \alpha_r     \right)_{l=1,\dots, K-1} \end{array} \right).
	\end{align*}
	\label{Lemma:grad}
\end{Theorem}

\noindent \textit{Proof}. The Riemannian gradient of a product manifold is the Cartesian product of the individual expressions \citep{Absil}. We compute the Riemannian gradients with respect to $S_1, \dots, S_K$ and $\eta$.\\

The gradient with respect to $\eta$ is the classical Euclidean gradient, hence we get by using the chain rule
\begin{align*}
\left(\grad \hat{\mathcal{L}}(\theta)\right)_{\eta_r} = \sum\limits_{i=1}^m\left( \sum\limits_{k=1}^K h^i(\theta_k)\right)^{-1} \sum\limits_{j=1}^K \frac{\partial h^i(\theta_j)}{\partial \eta_l} \sum\limits_{i=1}^m \sum\limits_{j=1}^K f_j^i \left( \one_{\{j=r\}} - \frac{\exp(\eta_r)}{\sum\limits_{k=1}^K \exp(\eta_k)} \right) ,
\end{align*}
	for $r=1, \dots, K-1$, where $\one_{\{j=r\}} =1$ if $j=r$ and $0$, else. \\
For the derivative of the penalizer with respect to $\eta_r$, we get
\begin{align*}
\left(\grad \Pen(\theta)\right)_{\eta_r} &=\left(\grad^{e} \Pen(\theta)\right)_{\eta_r} = \zeta \left(1-K\frac{\exp(\eta_r)}{\sum\limits_{j=1}^{K}\exp(\eta_j)}\right).
\end{align*}

The Riemannian gradient with respect to the matrices $S_1, \dots, S_K$ is the projected Euclidean gradient onto the subspace $T_{S_j} \bP^{d+1}$ (with inner product \eqref{inner_intr}), see \cite{Absil,boumal2020intromanifolds}. The relationship between the Euclidean gradient $\grad^{e} f$ and the Riemannian gradient $\grad f$ for an arbitrary function $f: \bP^n \rightarrow \R$ with respect to the intrinsic inner product defined on the set of positive definite matrices \eqref{inner_intr} reads
\begin{align}
\grad f(S) = \frac{1}{2} S \left( \grad^{e} f(S) +  \left(\grad^{e} f(S)\right)^T \right)S,
\label{Proof_grad:rel}
\end{align}
see for example \cite{Hosseini15,Jeuris}. In a first step, we thus compute the Euclidean gradient with respect to a matrix $S_l$:
\begin{align}
\left(\grad^{e} \hat{\mathcal{L}}(\theta)\right)_{S_l} = -\frac{1}{2} \sum\limits_{i=1}^m f_l^i (S_l^{-1}y_i {y_i}^T S_l^{-1}- S_l^{-1}),
\label{Proof_grad:mat}
\end{align}
where we used the Leibniz rule and the partial matrix derivatives
\begin{alignat*}{2}
\frac{\partial \left(\det(S_l)^{-1/2}\right)}{\partial S_l} &= -\frac{1}{2} (\det(S_l))^{-1/2} S_l^{-1}, \\
\frac{\partial \exp\left(-\frac{1}{2}y_i^TS_l^{-1}y_i\right)}{\partial S_l} &= \frac{1}{2}\exp\left(-\frac{1}{2}y_i^TS_l^{-1}y_i\right) S_l^{-1}y_i {y_i}^T S_l^{-1}, 
\end{alignat*}
which holds by the chain rule and the fact that $S_l^{-1}$ is symmetric.\\
Using the relationship  \eqref{Proof_grad:rel} and using \eqref{Proof_grad:mat} yields the Riemannian gradient with respect to $S_l$. It is given by
\begin{align*}
\left(\grad \hat{\mathcal{L}}(\theta)\right)_{S_l} =  \frac{1}{2} \sum\limits_{i=1}^m f_l^i ( y_i {y_i}^T - S_l).
\end{align*}
Analogously, we compute the Euclidean gradient of the matrix penalizer $\psi(S_j, \Phi)$ and use the relationship \eqref{Proof_grad:rel} to get the Riemannian gradient of the matrix penalizer.
\qed
\\

To apply Newton-like algorithms, we derived a formula for the Riemannian Hessian of the reformulated problem. It is stated in Theorem \ref{Lemma:Hess}:

\begin{Theorem}
	\label{Lemma:Hess}
	Let $\theta \in \Mf$ and $\xi_{\theta} \in T_{\theta}\Mf$.	The Riemannian Hessian is given by 
	\begin{align*}
	\Hess \left(\hat{\mathcal{L}}(\theta) \right)[\xi_{\theta}] = \left( \zeta_{S}, \zeta_{\eta}\right) \in T_{\theta}\Mf,
	\end{align*}
	where
	\begin{align}
	\zeta_{S_l} &= -\frac{1}{4} \sum\limits_{i=1}^m  {f_l }^i \bigg[ {C_l}^i -  \bigg( {a_l}^i- \sum\limits_{j=1}^K {f_j}^i{a_j}^i\bigg)(y_i {y_i}^T - S_l)  \bigg] \label{Hess1} \\
	\zeta_{\eta_r} &=  \frac{1}{2}  \sum\limits_{i=1}^m \bigg[ {f_r}^i \bigg({a_r}^i - \sum\limits_{j=1}^K  {f_j}^i {a_j}^i\bigg) - 2\alpha_r \left(\xi_{\eta_r} - \sum\limits_{j=1}^{K-1} \alpha_j \xi_{\eta_j}\right) \bigg]\label{Hess2}
	\end{align}
	for $l=1,\dots, K$, $r=1,\dots, K-1 $ and 
	\begin{align*}
	{a_l}^i &= {y_i}^T {S_l}^{-1} \xi_{S_l}{S_l}^{-1}y_i - \tr(S_l^{-1}\xi_{S_l}) + 2\xi_{\eta_l}, 
	&&{f_l}^i = \frac{h^i(\theta_l)}{\sum\limits_{j=1}^K h^i(\theta_j)}, \\
	{C_l}^i &= y_i {y_i}^T{S_l}^{-1}\xi_{S_l} + \xi_{S_l}{S_l}^{-1}y_i {y_i}^T, 
	&&\alpha_r =  \frac{\exp(\eta_r)}{\sum\limits_{k=1}^K \exp(\eta_k)},  \\
	\xi_{\eta_K} &\equiv 0.
	\end{align*}

	\noindent The Hessian for the additive penalizer reads $\Hess(\Pen(\theta))[\xi_{\theta}] = \left( {\zeta_{S}}^{pen}, {\zeta_{\eta}}^{pen} \right)$, where	
	\begin{align*}
	\zeta_{S_l}^{pen} = \frac{\beta}{4}\left(\Psi {S_l}^{-1} \xi_{S_l} + \xi_{S_l}{S_l}^{-1}\Psi\right), \quad 
	\zeta_{\eta_l}^{pen} = K\zeta \alpha_r \left( \xi_{\eta_r} - \sum\limits_{j=1}^{K-1} \alpha_j \xi_{\eta_j}\right).
	\end{align*}
\end{Theorem}

\noindent \textit{Proof}. It can be shown that for a product manifold $\mathcal{M} = \mathcal{M}_1 \times \mathcal{M}_2$ with $\nabla^1, \nabla^2$ being the Riemannian connections of $\mathcal{M}_1, \mathcal{M}_2$, respectively, the Riemannian connection $\nabla$ of $\mathcal{M}$ for $X, Y \in  \mathcal{M}$ is given by
\begin{align*}
\nabla_Y (X) = \nabla_{Y_1}^1 X_1  \times \nabla_{Y_2}^2 X_2,
\end{align*}
where $X_1, Y_1 \in T\mathcal{M}_1$ and $X_2, Y_2 \in T\mathcal{M}_2$ \citep{Carmo}.

Applying this to our problem, we apply the Riemannian connections of the single parts on the Riemannian gradient derived in Theorem \ref{Lemma:grad}. It reads
\begin{align}
\Hess \hat{\mathcal{L}}(\theta)[\xi_{\theta}] =  \nabla_{\theta} \grad \hat{\mathcal{L}}(\theta)	=  \left(\left(\nabla_{\xi_{S_l}} ^{(pd)} \grad \hat{\mathcal{L}}(\theta)\right)_{l}, \left(\nabla_{\xi_{\eta_r}}^{e} \grad \hat{\mathcal{L}}(\theta)\right)_{r}\right)^{T} 
\label{Proof:Hess}
\end{align}
for $l=1, \dots, K$ and $r=1,\dots, K-1$.\\ 
We will now specify the single components of \eqref{Proof:Hess}. Let 
\begin{align*}
\grad_{S_l}\hat{\mathcal{L}}(\theta) = \left(\grad \hat{\mathcal{L}}(\theta)\right)_{S_l},  \quad 
\grad_{\eta_r}\hat{\mathcal{L}}(\theta) = \left(\grad \hat{\mathcal{L}}(\theta)\right)_{\eta_r}
\end{align*}
denote the Riemannian gradient from Theorem \ref{Lemma:grad} at position $S_l$, $\eta_r$, respectively.

For the latter part in \eqref{Proof:Hess}, we observe that the Riemannian connection $\nabla_{\xi_{\eta_l}}^{e}$ for $\xi_{\eta_l} \in \R$ is the classical vector field differentiation \citep[Section 5.3]{Absil}. We obtain
\begin{align}
\nabla_{\xi_{\eta_l}}^{e}\grad \hat{\mathcal{L}}(\theta) &= \sum\limits_{j=1}^K \Deriv_{S_j} (\grad_{\eta_l}\hat{\mathcal{L}}(\theta))[\xi_{S_j}]  + \sum\limits_{j=1}^{K-1} \Deriv_{\eta_r} (\grad_{\eta_j} \hat{\mathcal{L}}(\theta))[\xi_{\eta_j}],
\label{Proof:Hess_eta1}
\end{align}
where $\Deriv_{S_j}(\cdot)[\xi_{S_j}]$, $\Deriv_{\eta_r}(\cdot)[\xi_{\eta_r}]$ denote the classical Fréchet derivatives with respect to $S_j$, $\eta_j$ along the directions $\xi_{S_j}$ and $\xi_{\eta_j}$, respectively.\\
For the first part on the right hand side of \eqref{Proof:Hess_eta1}, we have
\begin{align*}
\sum\limits_{j=1}^K \Deriv_{S_j}(\grad_{\eta_r} \hat{\mathcal{L}}(\theta))[\xi_{S_j}] &=  \frac{1}{2}\sum\limits_{i=1}^m \bigg[ \frac{h^i(\theta_r)}{\sum\limits_{k=1}^K h^i(\theta_k)}\bigg({y_i}^T {S_r}^{-1} \xi_{S_r}{S_r}^{-1}y_i - \tr(S_r^{-1}\xi_{S_r}) \\ 
&  \qquad   - \sum\limits_{j=1}^K \frac{h^i(\theta_j)}{\sum\limits_{k=1}^K h^i(\theta_j)} ({y_i}^T {S_j}^{-1} \xi_{S_j}{S_j}^{-1}y_i - \tr(S_j^{-1}\xi_{S_j}))\bigg)\bigg]
\end{align*}
and for the second part
\begin{align*}
\sum\limits_{j=1}^{K-1} \Deriv_{\eta_j}(\grad_{\eta_r} \hat{\mathcal{L}}(\theta))[\xi_{\eta_j}] &= \sum\limits_{i=1}^m \bigg[ \bigg(\frac{h^i(\theta_r)}{\sum\limits_{k=1}^K h^i(\theta_k)}-\alpha_r \bigg) \xi_{\eta_r}  +  \alpha_r \sum\limits_{j=1}^{K-1} \alpha_j \xi_{\eta_j}\\
& \qquad \qquad \qquad \qquad \qquad \quad - \frac{h^i(\theta_r)}{\sum\limits_{k=1}^K h^i(\theta_j)} \sum\limits_{j=1}^{K-1} \frac{h^i(\theta_j)}{\sum\limits_{k=1}^K h^i(\theta_j)} \xi_{\eta_j} \bigg]
\end{align*}
by applying the chain rule, the Leibniz rule and the relationship $\alpha_l = \frac{\exp(\eta_l)}{\sum\limits_{k=1}^K \exp(\eta_k)}$. Plugging the terms into \eqref{Proof:Hess_eta1}, this yields the expression for $\zeta_{\eta_r}$ in \eqref{Hess2}.\\

For the Hessian with respect to the matrices $S_l$, we first need to specify the Riemannian connection with respect to the inner product \eqref{inner_intr}. It is uniquely determined as the solution to the Koszul formula \citep[Section 5.3]{Absil}, hence we need to find an affine connection that satisfies the formula. For a positive definite matrix $S$ and symmetric matrices $\zeta_{S}, \xi_{S}$ and $\Deriv(\xi_S)[\zeta_{S}]$, this solution is given by \citep{Jeuris,Sra_Conic}
\begin{align*}
\nabla_{\nu_{S}}^{(pd)} \xi_{S} = \Deriv(\xi_S)[\nu_{S}] - \frac{1}{2} (\nu_{S} S^{-1}\xi_{S} + \xi_{S}S^{-1} \nu_{S}),		
\end{align*}
where $\xi_S$, $\nu_S$ are vector fields on $\Mf$ and $\Deriv(\xi_S)[\nu_{S}]$ denotes the classical Fréchet derivative of $\xi_S$ along the direction $\nu_S$.
Hence, for the first part in \eqref{Proof:Hess}, we get
\begin{align}
& \left(\nabla_{\xi_{S_l}} ^{(pd)} \grad \hat{\mathcal{L}}(\theta)\right)_l = \bigg(\sum\limits_{j=1}^K \Deriv_{S_j}(\grad_{S_l} \hat{\mathcal{L}}(\theta))[\xi_{S_j}] + \sum\limits_{j=1}^{K-1} \Deriv_{\eta_j} (\grad_{S_l} \hat{\mathcal{L}}(\theta))[\xi_{\eta_j}] 				\notag
\\ & \qquad  \qquad \qquad \qquad \qquad \qquad  \qquad - \frac{1}{2} \left(\grad_{S_l} \hat{\mathcal{L}}(\theta)S_l^{-1}\xi_{S_l} + \xi_{S_l} S_l^{-1} \grad_{S_l}\hat{\mathcal{L}}(\theta)\right)\bigg)_l.
\label{Proof:Hess_S1}
\end{align}
After applying the chain rule and Leibniz rule, we obtain
\begin{align}
\sum\limits_{j=1}^K \Deriv_{S_j}(\grad_{S_l}\hat{\mathcal{L}}(\theta))[\xi_{S_j}] &= -\frac{1}{4} \sum\limits_{i=1}^m  f_l^i \bigg[2\xi_{S_l} - \bigg(({y_i}^T {S_l}^{-1} \xi_{S_l}{S_l}^{-1}y_i - \tr({S_l}^{-1}\xi_{S_l})) \notag
\\&  + \sum\limits_{j=1}^K f_j^i ( {y_i}^T {S_j}^{-1} \xi_{S_j}{S_j}^{-1}y_i - \tr({S_j}^{-1}\xi_{S_j}))\bigg)  (y_i {y_i}^T - S_l) \bigg]
\label{Proof:Hess_S2}
\end{align}
and 
\begin{align}
\sum\limits_{j=1}^{K-1} \Deriv_{\eta_j} (\grad_{S_l}\hat{\mathcal{L}}(\theta))[\xi_{\eta_j}] &=  \frac{1}{2} \sum\limits_{i=1}^m \left( \frac{h^i(\theta_l)}{\sum\limits_{k=1}^K h^i(\theta_j)} \bigg(\xi_{\eta_l} - \sum\limits_{j=1}^{K-1} \frac{h^i(\theta_j)}{\sum\limits_{k=1}^K h^i(\theta_k)} \xi_{\eta_k}\bigg)\right) \notag\\
& \hspace{7cm} \times (y_i {y_i}^T - S_l).
\label{Proof:Hess_S3}
\end{align}
We plug \eqref{Proof:Hess_S2}, \eqref{Proof:Hess_S3} into \eqref{Proof:Hess_S1} and use the Riemannian gradient at position $S_l$ for the last term in \eqref{Proof:Hess_S1}. After some rearrangement of terms, we obtain the expression for $\zeta_{S_l}$ in \eqref{Hess2}.\\

The computation of $\Hess(\Pen(\theta))[\xi_{\theta}]$ is analogous by replacing $\hat{\mathcal{L}}$ with $\varphi(\eta, \zeta)$ in \eqref{Proof:Hess_eta1} and with $\psi(S_l, \Phi)$ in \eqref{Proof:Hess_S1}.

\qed

%% file: Chapters/RNTR.tex
Equipped with the Riemannian gradient and the Riemannian Hessian, we are now in the position to apply Newton-type algorithms to our optimization problem. As studying positive-definiteness of the Riemannian Hessian from Theorem \ref{Lemma:Hess} is hard, we suggest to introduce some safeguarding strategy for the Newton method by applying a Riemannian Newton Trust-Region method.

\subsection{Riemannian Newton Trust-Region Method}
\label{subsec: RTR}

The Riemannian Newton Trust-Region Algorithm is the retraction-based generalization of the standard Trust-Region method \citep{Conn} on manifolds, where the quadratic subproblem uses the Hessian information for an objective function $f$ that we seek to minimize. Theory on the Riemannian Newton Trust-Region method can be found in detail in \cite{Absil}, we here state the Riemannian Newton Trust-Region method in Algorithm \ref{alg:RTR}. Furthermore, we will study both global and local convergence theory for our penalized problem.

\begin{algorithm}[h]
	\caption{Riemannian Trust-Region algorithm \cite{Absil}}
	\label{alg:RTR}
	\SetAlgoLined
	\normalsize 
	\KwIn{objective $f$ with Hessian $H$ , initial iterate $\theta^{0} \in \Mf$, initial TR-radius $\Delta_0$, maximal TR-radius $\bar{\Delta}$, rejection threshold $\rho' \in [0, 1/4)$, acceptance parameters $0 \leq \omega_1 \le \omega_2 \leq 1, \tau_1 \leq 1/4, \tau_2 > 1$, termination criteria}
	Set $t=0$;\\
	\While{termination criteria not fulfilled}{
		Obtain $s^{t}$ by (approximately) solving the TR-subproblem
		\begin{align*}
		 \min_{s \in T_{\theta^{t}}\Mf} \hat{m}_{\theta^{t}}(s) =   f(\theta^{t}) + \langle \grad f(\theta^{t}), s\rangle_{\theta^{t}} + \frac{1}{2}\langle H_{t}[s],s\rangle_{\theta^{t}} \text{ s.t. }  \norm{s}_{\theta^{t}} \leq \Delta_t ;
		\end{align*} \label{subproblem} \\
		Evaluate $\rho_t = \frac{f(\theta^{t}) - f(R_{\theta^{t}}(s^{t}))}{\hat{m}_{\theta^{t}}(0_{\theta^{t}}) - \hat{m}_{\theta^{t}}(s^{t}) }$\\
		\uIf{$\rho_t < \omega_1$}{
			$\Delta_{t+1} = \tau_1 \Delta_t$; \label{reduce_delta}
		}\uElseIf{$\rho_t > \omega_2$ and $\norm{s^{t}}_{\theta^{t}} = \Delta_t$}{
			$\Delta_{t+1} = \min(\tau_2\Delta_t, \bar{\Delta})$;
		}\uElse{
			$\Delta_{t+1} = \Delta_t$;
		}
		\uIf{$\rho_t > \rho'$}{
			$\theta^{t+1} = R_{\theta^{t}}(s^{t})$;
		}\uElse{
			$\theta^{t+1} = \theta^{t}$
		}
		set $t=t+1$;
	}
	
\end{algorithm}

\paragraph{Global convergence.} In the following Theorem, we show that the Riemannian Newton Trust-Region Algorithm applied on the reformulated penalized problem converges to a stationary point under suitable conditions: we assume that the mixing proportion of each cluster is above a certain threshold in each iteration and that the covariance matrices $S_j$ do not get arbitrarily large in each iteration.

\begin{Theorem}(Global convergence)
	\label{Th:global_conv_Lhatpen}
	Consider the penalized reformulated objective $\hat{\mathcal{L}}_{pen}$ from \eqref{problem_w_pen}. Assume there exists $\epsilon > 0$ and $C > 0$ such that 
	\begin{align}
		\alpha_j^t = \frac{\exp(\eta_j^t)}{\sum\limits_{k=1}^K \exp(\eta_k^t)} > \epsilon
		\label{alpha_bound}
	\end{align}
	and that there exists $0<\tau^t <C$ such that
	\begin{align}
		\norm{S_j^t} \leq \tau^t \norm{\Psi}
		\label{S_jbounded}
	\end{align}
	for all iterations $t=0,1,\dots$. \\
	 Then, if we apply the Riemannian Newton Trust-Region Algorithm (Alg. \ref{alg:RTR}) to minimize ${f= -\hat{\mathcal{L}}_{pen}}$ yielding iterates $\theta^t$, it holds
	\begin{align}
	\lim\limits_{t \rightarrow \infty} \grad \hat{\mathcal{L}}_{pen}(\theta^{t}) = 0.
	\label{globconv_to_stat}
	\end{align}
	
\end{Theorem}

\noindent \textit{Proof.} 
According to general global convergence results for Riemannian manifolds, convergence to a stationary point is given if the iterates remain bounded (see \cite[Proposition 7.4.5]{Absil} and proofs in \cite[Section 7.4]{Absil}. Since the rejection threshold in {Algorithm \ref{alg:RTR}} is nonnegative, i.e. $\rho' > 0$, we get
\begin{align}
\hat{\mathcal{L}}_{pen}(\theta^0) \leq \hat{\mathcal{L}}_{pen} (\theta^t)
		\label{GMM:eq:incr_Lhat}
\end{align}
		for all iterations $t=0,1,2, \dots$.

We show that the iterates $S_j^t$ remain in the interior of $\bP^{d+1}$. For this, assume there exists a subsequence ${\theta^{t_i}}$ with $\lambda_{\min}(S_j^{t_i}) \rightarrow 0$ as $t_i \rightarrow \infty$. By the proof of Theorem \ref{Prop: Boundedness}, this implies $\hat{\mathcal{L}}_{pen} (\theta^{t_i}) \xrightarrow[]{t_i \rightarrow \infty} {-\infty}$ which is a contradiction to \eqref{GMM:eq:incr_Lhat}. Thus, there exists a lower bound $C_l > 0$ such that
\begin{align}
C_l \leq \lambda_{\min}({S_j^t}).
\label{GMM:Pr:globconv_upperS}
\end{align}
For the upper bound, we consider the set of successful (unsuccessful) steps $\mathfrak{S}_t$ ($\mathfrak{F}_t$) generated by the algorithm until iteration $t$ given by
\begin{align*}
\mathfrak{S}_t = \{l \in \{0,1,\dots, t\}: \rho_l > \rho'\}, \qquad
\mathfrak{F}_t = \{l \in \{0,1,\dots, t\}: \rho_l \leq \rho'\}.
\end{align*}
Let 
\begin{align*}
\xi^t = \left(\left(\xi_{S_1}^t, \dots, \xi_{S_K}^t \right), {\xi_{\eta}}^t \right) \in T_{\theta}\Mf_{GMM}
\end{align*} 
be the tangent vector returned by solving the quadratic subproblem in line \ref{subproblem}, Algorithm \ref{alg:RTR} and $R_{S_j^t}(\xi_{S_j^t}) = S_j^t \exp\left((S_j^t)^{-1} \xi_{S_j^t}\right)$ be the retraction of $\xi^t$ at iteration $t$ with respect to $S_j^t$, see \eqref{Retraction_probl}. 
Due to the boundedness of the quadratic subproblem in \eqref{alg:RTR}, there exists $\tilde{\Delta} > 0$ such that $\norm{{\xi_{S_j}}^t} \leq \tilde{\Delta}$ for all $j=1, \dots, K$. 
We get
\begin{align*}
\norm{\xi_{S_j}^t} &= \norm{R_{S_j^t}(\xi_{S_j^t})} \one_{\{t \in \mathfrak{S}_t\}} + \norm{S_j^{t-1}} \one_{\{t \in \mathfrak{F}_t\}} \\
& \leq \norm{S_j^{t-1}} \left(\exp \left( \norm{\left(S_j^{t-1}\right)^{-1}} \norm{\xi_{S_j}^t} \right) \one_{\{t \in \mathfrak{S}_t\}} + \one_{\{t \in \mathfrak{F}_t\}} \right) \\
& \leq \norm{S_j^{t-1}} \left( \one_{\{t \in \mathfrak{F}_t\}} + \exp\left( \frac{\norm{\xi_{S_j^t}}}{\lambda_{\min}(S_j^{t-1})} \right) \one_{\{t \in \mathfrak{S}_t\}} \right) \\
& \leq \norm{S_j^{t-1}} \left(\one_{\{t \in \mathfrak{F}_t\}}  + \exp \left(\frac{\tilde{\Delta}}{C_l} \right) \one_{\{t \in \mathfrak{S}_t\}} \right)	\\
& \leq \tau^{t-1} \norm{\Psi} \left(\one_{\{t \in \mathfrak{F}_t\}}  + \exp \left(\frac{\tilde{\Delta}}{C_l} \right) \one_{\{t \in \mathfrak{S}_t\}} \right)	
\end{align*}
and from the assumption \eqref{S_jbounded} boundedness from above follows directly.\\	

\noindent We now show boundedness of 
\begin{align*}
\eta^t = (\eta_1^t, \dots, \eta_{K-1}^t) \in \R^{K-1}.
\end{align*}
By the inequality \eqref{GMM:eq:incr_Lhat}, we have
\begin{align*}
\hat{\mathcal{L}}_{pen} (\theta^0) \leq \hat{\mathcal{L}}_{pen} (\theta^t)  = \hat{\mathcal{L}}(\theta^t) + \sum\limits_{j=1}^K \psi(S_j^t, \Psi) + \varphi(\eta^t, \zeta).
\end{align*}
Due to \eqref{GMM:Pr:globconv_upperS}, there exists $\tilde{C} > 0$ such that 
\begin{align}
\hat{\mathcal{L}}_{pen} (\theta^0) \leq \hat{\mathcal{L}}_{pen} (\theta^t)  \leq \tilde{C} + \varphi(\eta^t, \zeta).
\label{GMM:eq:hatL_bound}
\end{align}
We will study $\varphi(\eta^t, \zeta)$ for $\eta^t$ at the boundary in the following. We distinguish between the following two cases:
\begin{enumerate}
	\item \label{GMM:eq:conv_case1} Assume there exists $j \in \{1, \dots, l\}$ such that $\eta_j^t \rightarrow \infty$.
	\item \label{GMM:eq:conv_case2} Assume there does not exist a $j \in \{1, \dots, K-1\}$ such that $\eta_j^t \rightarrow \infty$.
\end{enumerate}
We first study the first case, that is \ref{GMM:eq:conv_case1} For this, we distinguish two cases:
\begin{itemize}
	\item[\ref{GMM:eq:conv_case1} a)] 	 	
	We study the case where only some of the $\eta_j^t$ approach $\infty$. Without loss of generality, assume that $\eta_j^t \rightarrow \infty$ for $j \in \{1, \dots, l\}$ for $l < K-1$. Further, for all $j > l$, assume that $\eta_j^t \leq c$ for a constant $c > 0$.  \\
	Recall that 
	\begin{align*}
	\alpha_j^t = \frac{\exp(\eta_j^t)}{\sum\limits_{k=1}^K \exp(\eta_k^t)}
	\end{align*}
	for all $j=1, \dots, K$ with $\eta_K = 0$. By assumption \eqref{alpha_bound}, we have $\alpha_j^t > \epsilon$. 	 	
	With the rule of L'H\^opital, we get
	\begin{align*}
	\lim\limits_{t \rightarrow \infty} \alpha_j^t = \lim\limits_{t \rightarrow \infty} \frac{\exp(\eta_j^t)}{\sum\limits_{j=1}^K \exp(\eta_k^t)} = \frac{1}{l},
	\end{align*}
	yielding
	\begin{align*}
	\lim\limits_{t \rightarrow \infty} \sum\limits_{j=1}^K \alpha_j^t = 1 + \sum \limits_{j=l+1}^K \lim\limits_{t \rightarrow \infty} \alpha_j^t \geq 1 + \sum\limits_{j=l+1}^K \epsilon_j  > 1 
	\end{align*}
	which is a contradiction to $\sum\limits_{j=1}^K \alpha_j = 1$.
	\item[\ref{GMM:eq:conv_case1} b)] Now assume that for all $j=1, \dots, K-1$
	we have $\eta_j^t \rightarrow \infty$ with the same speed, otherwise we are in case \ref{GMM:eq:conv_case1}a).  Without loss of generality, we set $n = \eta_j^t$ and let $n \rightarrow \infty$.
	Using $\eta_K=0$ and $\exp(n) > -1$, the penalization term $\varphi(\eta^t, \zeta)$ reads
	\begin{align*}
	\varphi(\eta^t, \zeta) &= \zeta \left(\sum\limits_{j=1}^{K} \eta_j^t - K \log \left(\sum\limits_{k=1}^K \exp(\eta_k^t)\right)\right) \\
	&= \zeta \left((K-1) n - K \log \left(1 + \sum\limits_{k=1}^{K-1} \exp(n) \right)\right) \\
	& \leq \zeta \left((K-1)n - K \log(K-2)-K\exp(n) \right)\\
	&= \zeta(-n-K\log(K-2)) \xrightarrow[n \rightarrow \infty]{} - \infty
	\end{align*}
	which is a contradiction to \eqref{GMM:eq:hatL_bound}.
\end{itemize}
We now study case \ref{GMM:eq:conv_case2}, that is we assume $\nexists j \in \{1, \dots, K-1\}: \eta_j^t \rightarrow \infty$. For this, assume $\exists l \leq K-1: j \in \{1, \dots, l\}: \eta_j^t \rightarrow -  \infty$ and for $j > l:$ $\vert \eta_j^t  \vert \leq c$. Then, the penalization term reads
\begin{align*}
\varphi(\eta^t, \zeta) &= \zeta \left(\sum\limits_{j=1}^{l} \eta_j^t + \sum\limits_{j=l+1}^{K-1} \eta_j^t - K \log \left(\sum\limits_{k=1}^K \exp(\eta_k^t)\right)\right) \\
&\leq \zeta \left(\sum\limits_{j=1}^l \eta_j^t + (K-1-l)c \right) \xrightarrow[\eta_j^t \rightarrow - \infty]{} - \infty,
\end{align*}
where we used $\eta_K=0$. As before, this is a contradiction to \eqref{GMM:eq:hatL_bound}.\\

Thus, the iterates $\{\theta\}^t$ remain in a compact set which yields the superlinear local convergence.	\qed	\\

We showed that Algorithm \ref{alg:RTR} applied on the problem of fitting Gaussian Mixture Models converges to a stationary point for all starting values. From general convergence theory for Riemannian Trust-Region algorithms \cite[Section 7.4.2]{Absil}, under some assumptions, the convergence speed of Algorithm \ref{alg:RTR} is superlinear. In the following, we show that these assumptions are fulfilled and specify the convergence rate.

\paragraph{Local convergence.} Local convergence result on (Riemannian) Trust Region-Methods depend on the solver for the quadratic subproblem. If the quadratic subproblem in Algorithm \ref{alg:RTR} is (approximately) solved with sufficient decrease, the local convergence close to a maximizer of $\hat{\mathcal{L}}_{pen}$ is superlinear under mild assumptions. The truncated Conjugate Gradient method is a typical choice that returns a sufficient decrease, we suggest to use this matrix-free method for Gaussian Mixture Models. We state the Algorithm in Appendix \ref{app:tCG}. The local superlinear convergence is stated in Theorem \ref{Th: local convergence}.
\begin{Theorem}(Local convergence)	
	\label{Th: local convergence}
	Consider Algorithm \ref{alg:RTR}, where the quadratic subproblem is solved by the truncated Conjugate Gradient Method and ${f =- \hat{\mathcal{L}}_{pen}}$. Let $v \in \Mf$ be a nondegenerate local minimizer of $f$, $H$ the Hessian of the reformulated penalized problem from Theorem \ref{Lemma:Hess} and $\delta < 1$ the parameter of the termination criterion of tCG (Appendix \ref{app:tCG}, line \ref{alg:tCG_termi} in Algorithm \ref{alg:tCG}). Then there exists $c >1$ such that, for all sequences $\{\theta^t\}$ generated by Algorithm \ref{alg:RTR} converging to $v$, there exists $T >0$ such that for all $t > T$,
	\begin{align}
	\norm{\grad \hat{\mathcal{L}}_{pen}(\theta^{t+1})} \leq c \norm{\grad \hat{\mathcal{L}}_{pen}(\theta^t)}^{\delta+1}. 
	\label{Th_loc_conv_eq}
	\end{align}  
\end{Theorem}

\noindent \textit{Proof.} Let $v \in \Mf$ be a nondegenerate local minimizer of $f$ (maximizer of $\hat{\mathcal{L}}_{pen}$). We choose the termination criterion of tCG such that $\delta <1$ . According to  \cite[Theorem 7.4.12]{Absil}, it suffices to show that $\Hess \hat{\mathcal{L}}_{pen}(R_{\theta}(\xi))$ is Lipschitz continuous at $0_{\theta}$ in a neighborhood of $v$. From the proof of Theorem \ref{Prop: Boundedness}, we know that close to $v$, we are bounded away from the boundary of $\Mf$, i.e. we are bounded away from points on the manifold with singular $S_j$, $j=1,\dots, K$. The extreme value theorem yields the local Lipschitz continuity such that all requirements of \cite[Theorem 7.4.12]{Absil} are fulfilled and the statement in Theorem \ref{Th: local convergence} holds true.
\qed

\subsection{Practical Design choices} 
We apply Algorithm \ref{alg:RTR} on our cost function \eqref{problem_w_pen}, where we seek to minimize \\$f= - \hat{\mathcal{L}}_{pen}$. The quadratic subproblem in line 3 in Algorithm \ref{alg:RTR} is solved by the truncated Conjugate Gradient method (tCG) with the inner product \eqref{inner_probl}. To speed up convergence of the tCG, we further use a preconditioner: At iteration $t$, we store the gradients computed in tCG and store an inverse Hessian approximation via the LBFGS formula. This inverse Hessian approximation is then used for the minimization of the next subproblem $\hat{m}_{\theta^{t+1}}$. The use of such preconditioners has been suggested by \cite{Morales_Prec} for solving a sequence of slowly varying systems of linear equations and gave a speed-up in convergence for our method.
The initial TR radius is set by using the method suggested by \cite{Sartenaer_initTR} that is based on the model trust along the steepest-descent direction. The choice of parameters $\omega_1, \omega_2, \tau_1, \tau_2, \rho'$ in Algorithm \ref{alg:RTR} are chosen according to the suggestions in \cite{Sens_TR_param} and \cite{Conn}.

%% file: Chapters/NumericalResults.tex
We test our method on the penalized objective function \eqref{opt_ref} on both simulated and real-world data sets. We compare our method against the (penalized) Expectation Maximization Algorithm and the Riemannian LBFGS method proposed by \cite{Hosseini15}. For all methods, we used the same initialization by running \textit{k-means++} \citep{Arthur} and stopped all methods when the difference of average log-likelihood for two subsequent iterates falls below $1e-10$ or when the number of iterations exceeds $1500$ (clustering) or $3000$ (density approximation). For the Riemannian LBFGS method suggested by \cite{Hosseini15,Hosseini20}, we used the MixEst package \citep{Mixest} kindly provided by one of the authors.\footnote{We also tested the Riemannian SGD method, but the runtimes turned out to be very high, so we omit the results here.} For the Riemannian Trust region method we mainly followed the code provided by the \textit{pymanopt} Python package provided by \cite{Pymanopt}, but adapted it for a faster implementation by computing the matrix inverse $S_j^{-1}$ only once per iteration. We used Python version 3.7. The experiments were conducted on an Intel Xeon CPU X5650 at 2.67 GHhz with 24 cores and 20GB RAM.\\

In Subsection \ref{subsec: 4.1}, we test our method on clustering problems on simulated data and on real-world datasets from the UCI Machine Learning Repository \citep{UCI_ML}. In Subsection \ref{subsec: 4.2} we consider Gaussian Mixture Models as probability density estimators and show the applicability of our method. 

\subsection{Clustering}
\label{subsec: 4.1}
We test our method for clustering tasks on different artificial (Subsection \ref{subsec: SimData}) and real-world (Subsection \ref{subsec: RealData}) data sets.

\subsubsection{Simulated Data}
\label{subsec: SimData}

As the convergence speed of EM depends on the level of separation between the data, we test our methods on data sets with different degrees of separation as proposed in \cite{DasGupta,Hosseini15}. The distributions are sampled such that their means satisfy
\begin{align*}
	\norm{\mu_{i} - \mu_{j}}^2 \geq c\max_{i,j}\left(\tr(\Sigma_{i}), \tr(\Sigma_{j})\right)
\end{align*}
for $i,j=1,\dots, K, i \neq  j$ and $c$ models the degree of separation. Additionally, a low \textit{eccentricity} (or condition number) of the covariance matrices has an impact on the performance of Expectation Maximization \citep{DasGupta}, for which reason we also consider different values of eccentricity $e = \sqrt{\left(\frac{\lambda_{max}(\Sigma_j)}{\lambda_{min}(\Sigma_j)}\right)}$. This is a measure of how much the data scatters.

We test our method on $20$ and $40$-dimensional data and an equal distribution among the clusters, i.e. we set $\alpha_j = \frac{1}{K}$ for all $j=1, \dots, K$. Although it is known that unbalanced mixing coefficients $\alpha_j$ also result in slower EM convergence, this effect is less strong than the level of overlap \citep{Naim}, so we only show simulation results with balanced clusters. Results for unbalanced mixing coefficients and varying number of components $K$ are shown for real-world data in Subsection \ref{subsec: RealData}.\\

\begin{figure}[h]
	\begin{subfigure}[b]{0.52\textwidth}	
		\includegraphics[width=\textwidth]{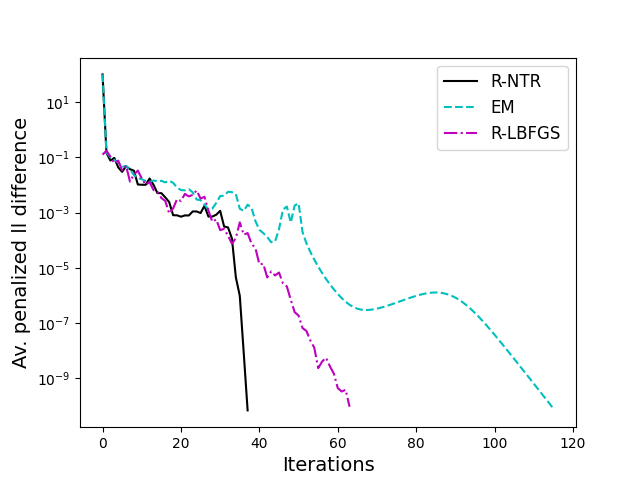}
		\caption{Average penalized log-likelihood red.} 
		\label{fig: ALL_red_lowsep}
	\end{subfigure}
	\hfill
	\begin{subfigure}[b]{0.47\textwidth}	
		\includegraphics[width=\textwidth]{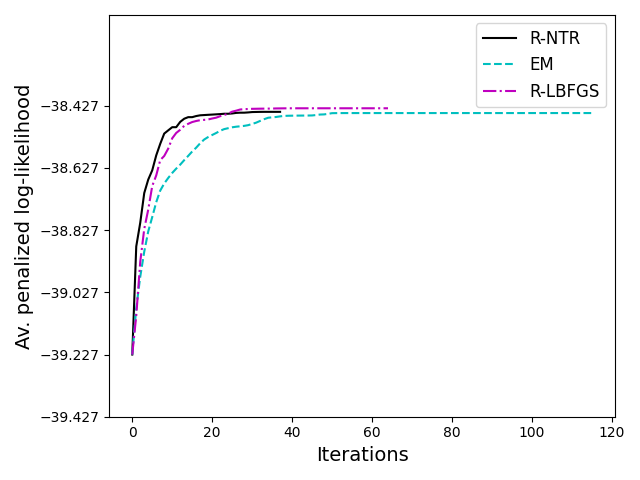}
		\caption{Average penalized log-likelihood} 
		\label{fig: ALL_lowsep}
	\end{subfigure}
	\caption{Average penalized log-likelihood reduction (a) and average penalized log-likelihood (b) for highly overlapping clusters: $d=20$, $K=5$, $e=1$, $c=0.2$.}
\end{figure}

	We first take a look at the $20$-dimensional data sets, for which we simulated $m=1000$ data points for each parameter setting. 
In Table \ref{tab:e=1_d20_K5}, we show the results for very scattered data, that is $e=1$. We see that, like predicted by literature, the Expectation Maximization converges slowly in such a case. This effect is even stronger with a lower separation constant $c$. The effect of the eccentricity becomes even more clear when comparing the results of Table \ref{tab:e=1_d20_K5} with Table \ref{tab:e=10,d20,K5}. Also the Riemannian algorithms converge slower for lower values of eccentricity $e$ and separation levels $c$. However, they seem to suffer less from hidden information than Expectation Maximization. The proposed Riemannian Newton Trust Region algorithm (R-NTR) beats the other methods in terms of runtime and number of iterations (see Figure \ref{fig: ALL_red_lowsep}). The Riemannian LBFGS (R-LBFGS) method by \cite{Hosseini15} also shows faster convergence than EM, but the gain of second-order information available by the Riemannian Hessian is obvious. However, the R-LBFGS results created by the MixEst toolbox show long runtimes compared to the other methods. 
 	We see from Figure \ref{fig: ALL_lowsep} that the average penalized log-likelihood is slightly higher for R-LBFGS in some experiments. Still, the objective evaluated at the point satisfying the termination criterion is at a competitive level in all methods (see also Table \ref{tab:e=1_d20_K5}). \\
When increasing the eccentricity (Table \ref{tab:e=10,d20,K5}), we see that the Riemannian methods still converge faster than EM, but our method is not faster than EM. This is because EM benefits from very low per-iteration costs and the gain in number of iterations is less strong in this case. However, we see that the Riemannian Newton Trust-Region method is not substantially slower. Furthermore, the average log-likelihood values (ALL) are more or less equal in all methods, so we might assume that all methods stopped close to a similar optimum. This is also underlined by comparable mean squared errors (MSE) to the true parameters from which the input data has been sampled from. In average, Riemannian Newton Trust-Region gives the best results in terms of runtime and number of iterations.

\begin{table}[h]
	\centering
	\footnotesize
	\caption{Simulation results of $20$ runs for dimensions $d=20$, number of components $K=5$ and eccentricity $e=1$}		
\begin{tabular}{lllll}
	\toprule
	&         &       EM &    R-NTR & R-LBFGS \\
	\midrule
	c=0.2 & Iterations &      295 &     79.4 &   113.4 \\
	& Mean time (s) &    3.833 &    2.716 &  16.008 \\
	& Mean ALL &   -42.64 &   -42.63 &  -42.63 \\
	& MSE weights &  0.00139 &  0.00143 &   0.008 \\
	& MSE means &     0.14 &     0.14 &    0.13 \\
	& MSE cov &     2.27 &      2.5 &    2.28 \\
	c=1 & Iterations &      262 &     47.5 &   102.7 \\
	& Mean time (s) &     3.65 &    2.069 &  14.281 \\
	& Mean ALL &    -41.2 &   -41.21 &  -41.21 \\
	& MSE weights &  0.00909 &  0.00985 &   0.008 \\
	& MSE means &     0.23 &     0.22 &    0.24 \\
	& MSE cov &     0.67 &     0.56 &     0.7 \\
	c=5 & Iterations &    208.8 &     54.2 &    92.4 \\
	& Mean time (s) &    2.747 &    2.123 &  12.975 \\
	& Mean ALL &   -36.98 &   -36.98 &  -36.99 \\
	& MSE weights &  0.00264 &  0.00282 &   0.008 \\
	& MSE means &     0.15 &     0.17 &    0.16 \\
	& MSE cov &     9.81 &      7.1 &   10.19 \\
	\bottomrule
\end{tabular}

\label{tab:e=1_d20_K5}
\end{table}

\begin{table}[h]
		\centering
	\footnotesize
	\caption{Simulation results of $20$ runs for dimensions $d=20$, number of components $K=5$ and eccentricity $e=10$}
	\begin{tabular}{lllll}
		\toprule
		&         &     EM &  R-NTR & R-LBFGS \\
		\midrule
		c=0.2 & Iterations &   66.2 &     16 &    33.2 \\
		& Mean time (s) &  0.873 &  0.798 &   4.012 \\
		& Mean ALL & -60.06 & -60.06 &  -60.07 \\
		& MSE weights &  3e-05 &  3e-05 &   0.008 \\
		& MSE means &   0.07 &   0.07 &    0.07 \\
		& MSE cov &   0.31 &   0.23 &    0.31 \\
		c=1 & Iterations &   56.6 &   17.4 &      30 \\
		& Mean time (s) &  0.748 &  0.818 &   3.624 \\
		& Mean ALL & -62.82 & -62.82 &  -62.83 \\
		& MSE weights &  3e-05 &  3e-05 &   0.008 \\
		& MSE means &   0.09 &   0.09 &    0.09 \\
		& MSE cov &   0.17 &   0.16 &    0.17 \\
		c=5 & Iterations &   43.1 &   14.7 &      29 \\
		& Mean time (s) &  0.617 &  0.748 &   3.378 \\
		& Mean ALL & -61.04 & -61.04 &  -61.05 \\
		& MSE weights &  4e-05 &  4e-05 &   0.008 \\
		& MSE means &   0.08 &   0.08 &    0.08 \\
		& MSE cov &   0.13 &   0.14 &    0.13 \\
		\bottomrule
	\end{tabular}
	\label{tab:e=10,d20,K5}
\end{table}

\begin{figure}[h!]
	\begin{subfigure}[b]{0.49\textwidth}	
		\includegraphics[width=\textwidth]{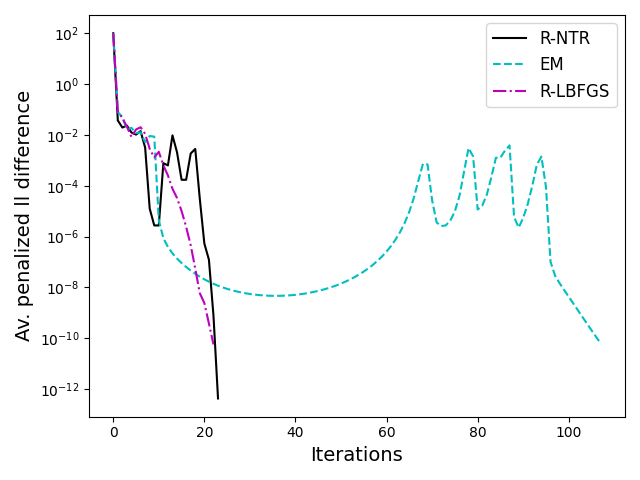}
		\caption{Average penalized log-likelihood red.} \label{GMM:fig:ALL_d=40}
		\label{fig: ALL_red_c5_d40}
	\end{subfigure}
	\hfill
	\begin{subfigure}[b]{0.49\textwidth}	
		\includegraphics[width=\textwidth]{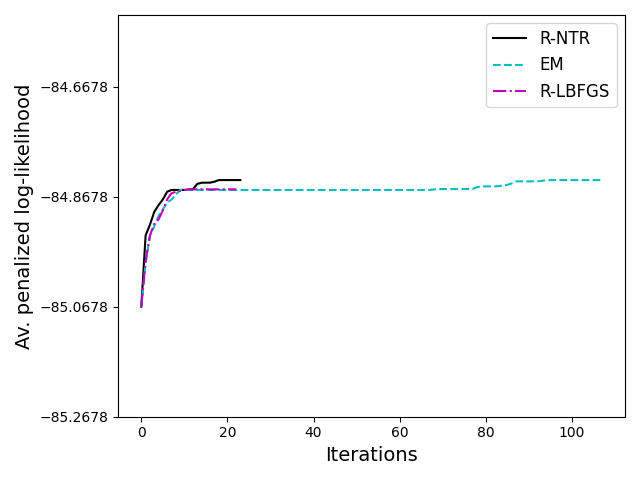}
		\caption{Average penalized log-likelihood} 
		\label{fig: ALLdiffL_d=40}
	\end{subfigure}
	\caption{Average penalized log-likelihood reduction (a) and average penalized log-likelihood (b) for overlapping clusters: $d=40$, $K=5$, $e=1$, $c=1$.}
\end{figure}

	In Table \eqref{tab:e=1,d40,K5,N1000}, we show results for dimension $d=40$ and low eccentricity ($e=1$) and the same simulation protocol as above (in particular, $m=1000$). We observed that with our method, we only performed very few Newton-like steps and instead exceeded the trust-region within the tCG many times, leading to poorer steps (see also Figure \ref{fig: ALLdiffL_d=40}). One possible reason is that the number of parameters increases with $d$ quadratically, that is in $\mathcal{O}(K d^2)$, while at the same time we did not increase the number of observations $m=1000$. If we are too far from a local optimum and the clusters are not well initialized due to few observations, the factor $f_l^i$ in the Hessian (Theorem \ref{Lemma:Hess}) becomes small, leading to large potential conjugate gradients steps (see Algorithm \ref{alg:tCG}). Although this affects the E-step in the Expectation Maximization algorithm as well, the effect seems to be much severe in our method.\\
	To underline this, we show simulation results for a higher number of observations, that is, $m=10.000$, in Table \ref{tab:e=1,d40,K5,N10000} with the same true parameters $\alpha_j, \mu_j, \Sigma_j$ as in Table \ref{tab:e=1,d40,K5,N1000}. As expected, the superiority in runtime of our method becomes visible: The R-NTR method beats Expectation maximization with a factor of $4$. Just like for the case of a lower dimension $d=20$, the mean average log-likelihood and the errors are comparable between our method and EM, whereas R-LBFGS shows slightly worse results although it now attains comparable runtimes to our method. \\
	
	We thus see that the ratio between number of observations and number of parameters must be large enough in order to benefit from the Hessian information in our method. 

\begin{table}[h]
		\centering
	\footnotesize
	\caption{Simulation results of $20$ runs for dimensions $d=40$, number of components $K=5$ and eccentricity $e=1$ with $m=1000$ observations}
\begin{tabular}{lllll}
		\toprule
		&         &       EM &    R-NTR &  R-LBFGS \\
		\midrule
		c=0.2 & Iterations &     57.4 &     27.9 &     40.7 \\
		& Mean time (s) &    1.332 &    2.296 &    6.744 \\
		& Mean ALL & -84.7935 &   -84.79 & -84.7895 \\
		& MSE weights &  0.00023 &  0.00023 &    0.008 \\
		& MSE means &    0.104 &     0.09 &    0.104 \\
		& MSE cov &    0.282 &    0.196 &    0.281 \\
		c=1 & Iterations &       61 &       29 &     48.6 \\
		& Mean time (s) &    1.391 &    2.476 &    8.282 \\
		& Mean ALL & -82.3395 & -82.3384 & -82.3358 \\
		& MSE weights &   0.0002 &   0.0002 &    0.008 \\
		& MSE means &    0.076 &    0.084 &    0.076 \\
		& MSE cov &    0.139 &    0.128 &     0.14 \\
		c=5 & Iterations &     81.8 &     28.8 &     49.2 \\
		& Mean time (s) &    1.812 &    2.203 &     8.62 \\
		& Mean ALL & -92.4886 & -92.4874 & -92.4925 \\
		& MSE weights &  0.00013 &  0.00013 &    0.008 \\
		& MSE means &     0.08 &    0.095 &     0.08 \\
		& MSE cov &    0.116 &    0.133 &    0.117 \\
		\bottomrule
	\end{tabular}
	\label{tab:e=1,d40,K5,N1000}
\end{table}

\begin{table}[h]
		\centering
	\footnotesize
	\caption{Simulation results of $20$ runs for dimensions $d=40$, number of components $K=5$ and eccentricity $e=1$ with $m=10000$ observations}
	\begin{tabular}{lllll}
			\toprule
			&         &       EM &    R-NTR &  R-LBFGS \\
			\midrule
			c=0.2 & Iterations &    350.4 &     33.2 &     69.4 \\
			& Mean time (s) &   53.621 &   12.455 &   15.449 \\
			& Mean ALL & -86.5717 & -86.5718 & -86.5731 \\
			& MSE weights &  0.00043 &  0.00043 &    0.008 \\
			& MSE means &    0.093 &    0.086 &    0.093 \\
			& MSE cov &    0.207 &    0.195 &    0.206 \\
			c=1 & Iterations &    495.6 &     63.6 &    107.3 \\
			& Mean time (s) &   79.955 &   20.739 &   23.153 \\
			& Mean ALL & -84.3783 & -84.3779 & -84.3797 \\
			& MSE weights &  0.00062 &  0.00065 &    0.008 \\
			& MSE means &    0.075 &    0.064 &    0.076 \\
			& MSE cov &    0.075 &    0.038 &    0.075 \\
			c=5 & Iterations &    260.4 &     28.8 &       54 \\
			& Mean time (s) &     42.6 &   10.434 &   11.692 \\
			& Mean ALL & -94.5592 & -94.5591 & -94.5603 \\
			& MSE weights &  0.00012 &  0.00013 &    0.008 \\
			& MSE means &    0.071 &    0.086 &    0.071 \\
			& MSE cov &    0.045 &    0.053 &    0.045 \\
			\bottomrule
		\end{tabular}
	\label{tab:e=1,d40,K5,N10000}
\end{table}

\subsubsection{Real-World Data}
\label{subsec: RealData}
We tested our method on some real-world data sets from UCI Machine Learning repository \citep{UCI_ML} besides the simulated data sets. For this, we normalized the data sets and tested the methods for different values of $K$.

\paragraph{Combined Cycle Power Plant Data Set \citep{Dataset_powerplant}.}
In Table \ref{tab:PowerPlant}, we show the results for the combined cycle power plant data set, which is a data set with $4$ features. Although the dimension is quite low, we see that we can beat EM both in terms of runtime and number of iterations for almost all $K$ by applying the Riemannian Newton-Trust Region method. This underlines the results shown for artificial data in Subsection \ref{subsec: SimData}. The gain by our method becomes even stronger when we consider a large number of components $K$ where the overlap between clusters is large and we can reach a local optimum with our method in up to $15$ times less iterations and a time saving of factor close to $4$.

\begin{table}[h]
		\centering
	\footnotesize
	\caption{Results of (normalized) combined cycle power plant data set for different number of components.Number of observations $m=9568$, dimensions $d=4$.}
	
	\begin{tabular}{llrrr}
	\toprule
	&     &       EM &  R-NTR &  R-LBFGS \\
	\midrule
	K = 2 & Time (s) &     0.40 &   0.63 &     2.38 \\
	& Iterations &     56 &  19 &    34 \\
	& ALL &    -4.24 &  -4.24 &    -4.24 \\
	K = 5 & Time (s) &     3.29 &   2.26 &     7.50 \\
	& Iterations &   239 &  48 &    70 \\
	& ALL &    -4.01 &  -4.01 &    -4.01 \\
	K = 10 & Time (s) &    31.72 &   4.28 &    23.40 \\
	& Iterations &  1097 &  58 &   110 \\
	& ALL &    -3.83 &  -3.82 &    -3.83 \\
	K = 15 & Time (s) &    28.27 &   6.79 &    35.77 \\
	& Iterations &   677 &  67 &   111 \\
	& ALL &    -3.75 &  -3.75 &    -3.75 \\
	\bottomrule
\end{tabular}
\label{tab:PowerPlant}
\end{table}

\paragraph{MAGIC Gamma Telescope Data Set \citep{Dataset_gammatelesc}.}
We also study the behaviour on a data set with higher dimensions and a larger number of observations with the MAGIC Gamma Telescope Data Set, see Table \ref{tab:GammaTelescope}. Here, we can also observe a lower number of iterations in the Riemannian Optimization methods. Similar to the combined cycle power plant data set, this effect becomes even stronger for a high number of clusters where the ratio of hidden information is large. Our method shows by far the best runtimes. For this data set, the average log-likelihood values are very close to each other except for $K=15$ where the ALL is worse for the Riemannian methods. It seems that in this case, the R-NTR and the R-LBFGS methods end in different local maxima than the EM. However, for all of the methods, convergence to global maxima is theoretically not ensured and for all methods, a globalization strategy like a split-and-merge approach \citep{Li_Li} might improve the final ALL values. As the Magic Gamma telescope data set is a classification data set with $2$ classes, we further report the classification performance in Table \ref{tab:GammaTelescope_wmse}. We see that the geodesic distance defined on the manifold and  the weighted mean squared errors (wMSE) are comparable between all three methods. In Table \ref{tab:gamma_ARI}, we also report the Adjusted Rand Index \citep{ARI} for all methods. Although the clustering performance is very low compared to the true class labels (first row), we see that it is equal among the three methods.\\

\begin{table}[h]
		\centering
	\footnotesize
	\caption{Results of (normalized) Magic Gamma telescope data set for different number of components. Number of observations $m=19020$, dimensions $d=11$.}
	\begin{tabular}{llrrr}
	\toprule
	&     &      EM &  R-NTR &  R-LBFGS \\
	\midrule
	K = 2 & Time (s) &    1.18 &   0.57 &     1.52 \\
	& Iterations &   30 &   6 &    17 \\
	& ALL &   -7.81 &  -7.81 &    -7.81 \\
	K = 5 & Time (s) &    3.96 &   1.32 &     5.37 \\
	& Iterations &   65 &   9 &    34 \\
	& ALL &   -6.53 &  -6.53 &    -6.53 \\
	K = 10 & Time (s) &   36.00 &   6.99 &    20.78 \\
	& Iterations &  293 &  34 &    77 \\
	& ALL &   -6.02 &  -6.02 &    -6.02 \\
	K = 15 & Time (s) &   56.24 &  12.61 &    50.52 \\
	& Iterations &  354 &  38 &   115 \\
	& ALL &   -5.39 &  -5.54 &    -5.51 \\
	\bottomrule
\end{tabular}
\label{tab:GammaTelescope}
\end{table}

\begin{table}[h]
	\footnotesize
	\caption{Model quality for (normalized) magic gamma telescope data set for $K=2$.}
	\begin{subtable}[t]{.51\textwidth}
		\centering
		\caption{geodesic distance and weighted MSE}
		\resizebox{!}{.15\textwidth}{
			\begin{tabular}{lrrr}
				\toprule
				{} &        EM &     R-NTR &   R-LBFGS \\
				\midrule
				distance    &  4.783910 &  4.783934 &  4.783922 \\
				wMSE weight &  0.000034 &  0.000034 &  0.000034 \\
				wMSE mean   &  1.883762 &  1.883765 &  1.883759 \\
				wMSE cov    &  8.214824 &  8.214840 &  8.214828 \\
				\bottomrule
		\end{tabular}}\label{tab:GammaTelescope_wmse}
	\end{subtable}
	\begin{subtable}[t]{.49\textwidth}
		\centering
		\caption{adjusted rand index }
		\resizebox{!}{.15\textwidth}{
			\begin{tabular}{llllr}
				\toprule
				{} & truth &    EM & R-NTR &  R-LBFGS \\
				\midrule
				truth   &     1 &  0.06 &  0.06 &     0.06 \\
				EM      &       &     1 &     1.00 &     1.00 \\
				R-NTR   &       &       &     1 &     1.00 \\
				R-LBFGS &       &       &       &     1 \\
				\bottomrule
		\end{tabular}}
		\label{tab:gamma_ARI}
	\end{subtable}
\end{table}


We show results on additional real-world data sets in Appendix \ref{app:data}.

\subsection{Gaussian Mixture Models as Density Approximaters}
\label{subsec: 4.2}
Besides the task of clustering (multivariate) data, Gaussian Mixture Models are also well-known to serve as probability density approximators of smooth density functions with enough components \citep{Scott_DensEst}.\\
In this Subsection, we present the applicability of our method for Gaussian mixture density approximation of a Beta-Gamma distribution and compare against EM and R-LBFGS for the estimation of the Gaussian components.\\

\begin{figure}	\begin{center}
		\includegraphics[scale=.5]{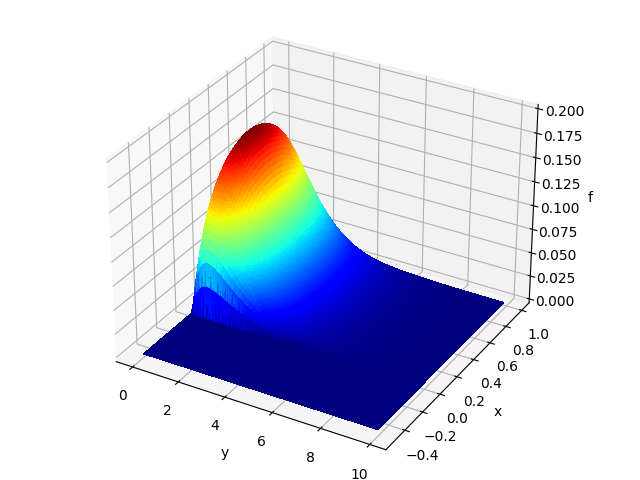}
		\caption{Probability density of bivariate 
				Beta($0.5$,$0.5$)-Gamma($1$, $1$) distribution.
			} 
		\label{fig: true_dens}\end{center}
\end{figure}

We consider a bivariate Beta-Gamma distribution with parameters $\alpha_{Beta} =0.5$, $\beta_{Beta} =0.5$, $a_{Gamma}=1,
\beta_{Gamma} = 1
$, where the joint distribution is characterized by a Gaussian copula. The density function surface is visualized in Figure \ref{fig: true_dens}.

\begin{figure}[!tbp]
	\begin{subfigure}[b]{0.49\textwidth}
		\includegraphics[width=\textwidth]{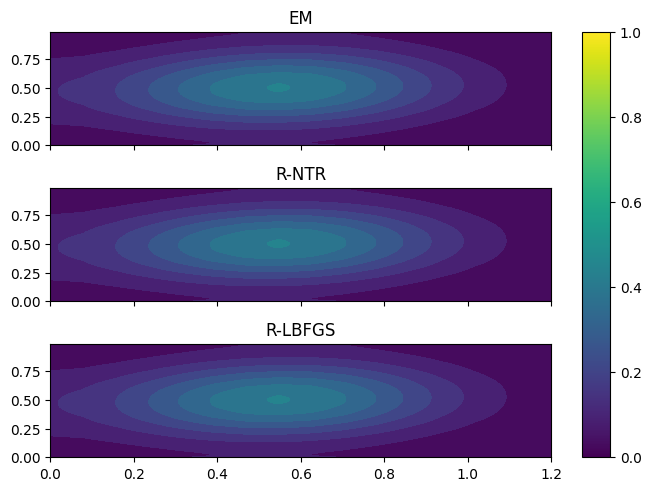}
		\caption{$K=2$}
		\label{fig:f1}
	\end{subfigure}
	\hfill
	\begin{subfigure}[b]{0.49\textwidth}
		\includegraphics[width=\textwidth]{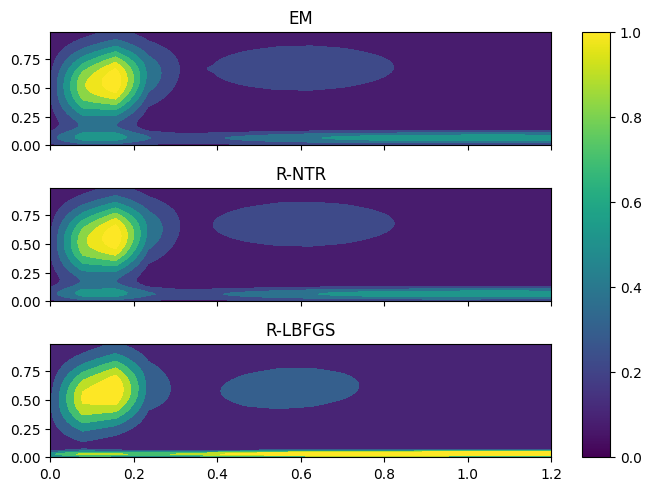}
		\caption{$K=5$}
		\label{fig:f2}
	\end{subfigure}
	\caption{Contours of pointwise root mean squared error (RMSE) 
			for density approximation via GMMs of the Beta($0.5$,$0.5$)-Gamma($1$, $1$) distribution.
		}
	\label{fig: RMSE}
\end{figure}

We simulated $1000$ realizations of the 
	Beta($0.5$,$0.5$)-Gamma($1$, $1$) distribution
 and fitted a Gaussian Mixture Model for $100$ simulation runs. We considered different numbers of $K$ and compared the (approximated) root mean integrated squared error (RMISE).\\
The RMISE and its computational approximation formula is given by \citep{Gross}:
\begin{align*}
RMISE(\hat{f}) &= \sqrt{\mathbb{E} \left(\int \left(f(x) - \hat{f}(x)\right)^2 dx\right)} \\
&\approx \sqrt{\frac{1}{N} \sum\limits_{r=1}^N \left(f(g_r) - \hat{f}(g_r)\right)^2 \delta_g^2 },
\end{align*}
where $f$ denotes the underlying true density , i.e.\\
	Beta($0.5$,$0.5$)-Gamma($1$, $1$),
$\hat{f}$ the density approximator (GMM), $N$ is the number of equidistant grid points with the associated grid width $\delta_g$.\\
For our simulation study, we chose $16384$ grid points in the box $[0,5] \times [0,10]$. We show the results in Table \ref{tab:RMISE}, where we fit the parameters of the GMM by our method (R-NTR) and compare against GMM approximations where we fit the parameters with EM and R-LBFGS. We observe that the RMISE is of comparable size for all methods and even slightly better for our method for $K=2$, $K=5$ and $K=10$. Just as for the clustering results in Subsection \ref{subsec: 4.1}, we have much lower runtimes for R-NTR and a much lower number of total iterations. This is a remarkable improvement especially for a larger number of components. We also observe that in all methods, the mean average log-likelihood (ALL) of the 
	training data sets with $1000$ observations attains higher values 
with an increasing number of components $K$. This supports the fact that the approximation power of GMMs for arbitrary density function is expected to become higher if we add additional Gaussian components \citep{Scott_DensEst,Goodfellow}. On the other hand, the RMISE (which is not based on the training data) increased in our experiments with larger $K$'s. This means that we are in a situation of overfitting. The drawback of overfitting is well-known for EM \citep{Andrews} and we also observed this for the R-NTR and the R-LBFGS methods. However, the RMISE are comparable and so none of the methods outperforms another substantially in terms of overfitting. This can also be seen from Figure \ref{fig: RMSE} showing the distribution of the pointwise errors for $K=2$ and $K=5$. Although the R-LBFGS method shows higher error values on the boundary of the support of the distribution for $K=5$, the errors show similar distributions among the three methods at a comparable level.
We propose methodologies such as cross validation \citep{murphy2013machine} or applying a split-and-merge approach on the optimized parameters \citep{Li_Li} to address the problem of overfitting.\\

\begin{table}[h]
		\centering
	\footnotesize
	\caption{Simulation results averaged over $100$ simulation runs for approximation of a 			Beta($0.5$,$0.5$)-Gamma($1$, $1$) distribution
		by a Gaussian Mixture Models with different values of $K$. Parameter Estimation by EM, R-NTR and R-LBFGS.}
	\begin{tabular}{lllll}
		\toprule
		&          &       EM &    R-NTR &  R-LBFGS \\
		\midrule
		K=2 & Mean RMISE &  0.00453 &  0.00453 &  0.00453 \\
		& SE RMISE &  0.00017 &  0.00017 &  0.00017 \\
		& Iterations &      113 &     27.9 &     25.9 \\
		& Mean time (s) &   0.2002 &   0.1532 &   0.9877 \\
		& Mean ALL & -1.53103 & -1.52953 & -1.53103 \\
		K=5 & Mean RMISE &  0.00565 &  0.00565 &  0.00567 \\
		& SE RMISE &  0.00028 &  0.00024 &  0.00025 \\
		& Iterations &    346.5 &       47 &     72.9 \\
		& Mean time (s) &   1.0893 &   0.5202 &   6.7941 \\
		& Mean ALL & -1.22852 & -1.22108 & -1.22912 \\
		K=10 & Mean RMISE &  0.00639 &  0.00643 &  0.00643 \\
		& SE RMISE &  0.00025 &  0.00027 &  0.00029 \\
		& Iterations &    623.1 &     56.8 &    112.2 \\
		& Mean time (s) &   3.4226 &   1.7108 &  20.4754 \\
		& Mean ALL & -1.06821 & -1.02844 & -1.06781 \\
		K=15 & Mean RMISE &  0.00667 &  0.00669 &   0.0067 \\
		& SE RMISE &  0.00026 &  0.00027 &  0.00026 \\
		& Iterations &    791.5 &     62.9 &    135.2 \\
		& Mean time (s) &   6.1649 &   2.5701 &     37.6 \\
		& Mean ALL & -1.02677 & -0.95907 & -1.02604 \\
		K=20 & Mean RMISE &  0.00681 &  0.00683 &  0.00685 \\
		& SE RMISE &  0.00026 &  0.00025 &  0.00026 \\
		& Iterations &    874.8 &     66.9 &    144.6 \\
		& Mean time (s) &   8.7694 &   3.3773 &  55.6337 \\
		& Mean ALL & -1.02029 &  -0.9202 & -1.02056 \\
		\bottomrule
	\end{tabular}
	
	\label{tab:RMISE}
\end{table}

The results show that our method is well-suited for both density estimation tasks and especially for clustering of real-world and simulated data.

%% file: Chapters/Conclusion.tex
 We proposed a Riemannian Newton Trust-Region method for Gaussian Mixture Models. For this, we derived an explicit formula for the Riemannian Hessian and gave results on convergence theory. Our method is a fast alternative to the well-known Expectation Maximization and existing Riemannian approaches for Gaussian Mixture Models. Especially for highly overlapping components, the numerical results show that our method leads to an enourmous speed-up both in terms of runtime and the total number of iterations and medium-sized problems. This makes it a favorable method for density approximation as well as for difficult clustering tasks for data with higher dimensions. Here, especially the availability of the Riemannian Hessian increases the convergence speed compared to Quasi-Newton algorithms. 
When considering higher-dimensional data, we experimentally observed that our method still works very well and is faster than EM if the number of observations is large enough. We plan to examine this further and take a look into data sets with higher dimensions. Here, it is common to impose a special structure on the covariance matrices to tackle the curse of dimensionality \citep{McLachlan}. Adapted versions for Expectation Maximization exist and a transfer of Riemannian method into constrained settings is subject of current research.

%% file: Chapters/Appendix/Unboundedness.tex
\begin{Theorem}
	The reformulated objective \eqref{opt_ref} without penalization term is unbounded from above.
\end{Theorem}
\textit{Proof.} We show that there exists $\theta_s$ such that $\lim\limits_{\theta \rightarrow \theta_s} \hat{\mathcal{L}} (\theta)= \infty$. \\ 
We consider the simplified case where $K=1$ and investigate the log-likelihood of the variable $\theta^{(n)} = S^{(n)}$ since $\alpha_1 = 1$.
Similar to the proof of Theorem \ref{Prop: Boundedness}, we introduce a singular positive semidefinite matrix $S_s$ with rank $r < d+1$ and a sequence of positive definite matrices $S_1^{(n)}$ converging to $S_s$ for $n \rightarrow \infty$. The reformulated objective at $\theta^{(n)}$ reads
\begin{align}
\hat{\mathcal{L}}\left(\theta^{(n)}\right) =
 \sum\limits_{i=1}^m \left(\log \left(\frac{1}{(2\pi)^{d/2}} \frac{\exp(\frac{1}{2}(1-{y_i}^T ({S_1}^{(n)})^{-1} y_i))}{\det\left(S_1^{(n)}\right)^{1/2}}  \right)\right). \label{Pr:appendix_Lhat}
\end{align}
With the decomposition $(S_1^{(n)})^{-1} = U^T D^n U$ as in \eqref{Proof:bound_dec}, we see that
\begin{align*}
y_i^T (S_1^{(n)})^{-1} y_i = \sum\limits_{l=1}^{d-r-1} \frac{\left((Uy_i)_l\right)^2}{\lambda_l} + \sum\limits_{l=d-r}^{d+1}\frac{\left((Uy_i)_l\right)^2}{\lambda_l}
\end{align*}
Now assume that one of the $y_i$ is in the kernel of the matrix $\tilde{U}_{s}$, where
\begin{align*}
\tilde{U}_{s} = \left(\begin{array}{ccc} u_{11} & \dots & u_{1(d-r-1)} \\ \vdots & \ddots & \vdots \\ u_{(d-r-1)1} & \dots & u_{(d-r-1)(d-r-1)}  \end{array}\right).
\end{align*}
Then we see that \eqref{Pr:appendix_Lhat} reads
\begin{align*}
\hat{\mathcal{L}}\left(\theta^{(n)}\right) = \sum\limits_{i=1}^m \bigg[\log \bigg(\frac{1}{(2\pi)^{d/2}} \frac{1}{\lambda^{(n)} \lambda^{+}}   \exp\bigg(\frac{1}{2}\bigg(1- \sum\limits_{l=r-d}^{d+1} \frac{\left((Uy_i)_l\right)^2}{\lambda_l} \bigg)\bigg)\bigg) \bigg] \underrightarrow{n \rightarrow \infty} \quad \infty,
\end{align*}
where $\lambda^{(n)} = \prod\limits_{1}^{d-r-1} \lambda_l^{(n)}$ and $\lambda^{+} = \prod\limits_{d-r}^{d+1} \lambda_l$ as in the proof of Theorem \ref{Prop: Boundedness}.

%% file: Chapters/Appendix/tCG.tex
We here state the truncated CG method to solve the quadratic subproblem in the Riemannian Newton Trust-Region Algorithm, line \ref{subproblem} in Algorithm \ref{alg:RTR}.

\begin{algorithm}
	\caption{Truncated Conjugate Gradient Method \cite{Absil}}	
	\label{alg:tCG}
	\SetAlgoLined
	\normalsize
	\KwIn{Riemannian gradient $\grad f(\theta^t)$, linear operator $H_t$, inner product $\langle \cdot,\cdot \rangle_{\theta^t}$, optional preconditioner $M_t$, TR radius $\Delta_t$, termination parameter $\delta$, $\kappa$}
	Set $s_0 = 0$, $r_0 = \grad f(\theta^t)$, $z_0 = M_t r_0$, $p_0 = -z_0$;\\
	\For{$n=0,1,2,\dots$}{
		\uIf{$\langle p_n, H_t[p_n]\rangle_{\theta^t} \leq 0$}{
			Compute $\tau^{tCG} = \arg\min\limits_{\tau} \hat{m}_{\theta^t}(s_n + \tau p_n)$ s.t. $\norm{s_n + \tau p_n}_{\theta^t} = \Delta_t$;\\
			Set $s_{n+1} = s_n + \tau^{tCG}p_n$;\\
			\textbf{return} $s_{n+1}$;
		}
	Compute ${\alpha_n^{tCG}} = \frac{\langle r_n, z_n \rangle_{\theta^t}}{\langle p_n, H_t[p_n]\rangle_{\theta_t}}$;\\
	\uIf{$\norm{s_n + \alpha_n^{tCG}p_n}_{\theta^t} > \Delta_t$}{
				Compute $\tau^{tCG} = \arg\min\limits_{\tau} \hat{m}_{\theta^t}(s_n + \tau p_n)$ s.t. $\norm{s_n + \tau p_n}_{\theta^t} = \Delta_t$;\\
			Set $s_{n+1} = s_n + \tau^{tCG}p_n$;\\
			\textbf{return} $s_{n+1}$;
		}\uElse{
			$s_{n+1} = s_n + \alpha_n^{tCG} p_n$;\\
			$r_{n+1} = r_n + \alpha_n^{tCG} H_t[p_n]$;\\
			$z_{n+1} = M_t r_{n+1}$;\\
			$\beta_{n+1}^{tCG} = \frac{\langle z_{n+1} z_{n+1} \rangle_{\theta^t}}{\langle z_n, z_n \rangle_{\theta^t}}$;\\
			$p_{n+1} = -z_{n+1} + \beta_{n+1}^{tCG} p_n$;\\
			\uIf{$\norm{r_{n+1}}_{\theta^t} \leq \norm{r_{0}}_{\theta^t} \min(\norm{r_{0}}_{\theta^t}^{\delta}, \kappa)$}{ \label{alg:tCG_termi}
			\textbf{return} $s_{n+1}$;}
		}		
	}
	
\end{algorithm}
\FloatBarrier

%% file: Chapters/Appendix/AdditionalNumerical.tex
\begin{table}[!hbt]
	\footnotesize
	\centering
	\caption{Simulation results of $20$ runs for dimensions $d=20$, number of components $K=5$ and eccentricity $e=5$ with $n=1000$ observations.}
	\begin{tabular}{lllll}
		\toprule
		&         &       EM &    R-NTR & R-LBFGS \\
		\midrule
		c=0.2 & Iterations &    145.7 &     38.5 &    60.4 \\
		& Mean time (s) &    1.899 &    1.512 &   8.165 \\
		& Mean ALL &   -50.65 &   -50.66 &  -50.66 \\
		& MSE weights &  0.00018 &  0.00019 &   0.008 \\
		& MSE means &     0.08 &     0.08 &    0.08 \\
		& MSE cov &     2.18 &     1.83 &    2.21 \\
		c=1 & Iterations &    258.9 &     60.4 &    84.1 \\
		& Mean time (s) &    3.061 &    1.951 &   11.64 \\
		& Mean ALL &   -57.65 &   -57.65 &  -57.65 \\
		& MSE weights &  0.00037 &   0.0004 &   0.008 \\
		& MSE means &     0.08 &     0.07 &    0.07 \\
		& MSE cov &     1.02 &     1.39 &    1.04 \\
		c=5 & Iterations &    194.6 &     37.2 &    65.7 \\
		& Mean time (s) &    2.367 &    1.267 &   8.927 \\
		& Mean ALL &   -54.86 &   -54.86 &  -54.87 \\
		& MSE weights &  0.00017 &  0.00018 &   0.008 \\
		& MSE means &     0.07 &     0.07 &    0.07 \\
		& MSE cov &     0.15 &     0.13 &    0.15 \\
		\bottomrule
	\end{tabular}
	
	\label{tab:e=5,d20,K5}
\end{table}

\begin{table}[h]
	\footnotesize
	\centering
	\caption{Simulation results of $20$ runs for dimensions $d=40$, number of components $K=5$ and eccentricity $e=10$ for different number of observations}
	
	\begin{subtable}[t]{.49\textwidth}
		\caption{$n=1000$ observations}
		\resizebox{!}{.44\textwidth}{
			
			\begin{tabular}{lllll}
				\toprule
				&         &       EM &    R-NTR &  R-LBFGS \\
				\midrule
				c=0.2 & Iterations &        4 &     16.3 &      6.6 \\
				& Mean time (s) &    0.091 &    1.336 &    0.622 \\
				& Mean ALL &  -125.63 &  -125.63 & -125.649 \\
				& MSE weights &    3e-05 &    3e-05 &    0.008 \\
				& MSE means &    0.091 &    0.095 &    0.091 \\
				& MSE cov &    0.224 &    0.215 &    0.224 \\
				\hline
				c=1 & Iterations &        4 &     17.6 &      8.8 \\
				& Mean time (s) &    0.097 &    1.561 &    0.862 \\
				& Mean ALL & -110.112 & -110.112 &  -110.14 \\
				& MSE weights &    3e-05 &    3e-05 &    0.008 \\
				& MSE means &    0.089 &    0.084 &    0.089 \\
				& MSE cov &    0.461 &    0.337 &    0.461 \\
				\hline
				c=5 & Iterations &        4 &       17 &      7.2 \\
				& Mean time (s) &    0.094 &    1.497 &    0.697 \\
				& Mean ALL & -116.276 & -116.276 &   -116.3 \\
				& MSE weights &    3e-05 &    3e-05 &    0.008 \\
				& MSE means &    0.061 &    0.066 &    0.061 \\
				& MSE cov &    0.351 &     0.43 &    0.351 \\
				\bottomrule
		\end{tabular} }
		\label{tab:e=10,d40,K5,N1000}
	\end{subtable}
	\begin{subtable}[t]{.44\textwidth}
		\caption{$n=10.000$ observations}
		\resizebox{!}{.49\textwidth}{
			\begin{tabular}{lllll}
				\toprule
				&         &       EM &    R-NTR &  R-LBFGS \\
				\midrule
				c=0.2 & Iterations &        4 &      2.3 &      3.3 \\
				& Mean time (s) &    0.628 &    1.524 &     0.74 \\
				& Mean ALL & -127.762 & -127.762 & -127.764 \\
				& MSE weights &        0 &        0 &    0.008 \\
				& MSE means &    0.089 &     0.09 &    0.089 \\
				& MSE cov &     0.14 &    0.189 &     0.14 \\
				\hline
				c=1 & Iterations &        4 &      2.8 &      3.8 \\
				& Mean time (s) &    0.609 &    1.676 &    0.884 \\
				& Mean ALL &  -112.25 &  -112.25 & -112.252 \\
				& MSE weights &        0 &        0 &    0.008 \\
				& MSE means &    0.084 &    0.084 &    0.084 \\
				& MSE cov &    0.313 &     0.29 &    0.313 \\
				\hline
				c=5 & Iterations &        4 &      2.6 &      3.6 \\
				& Mean time (s) &     0.63 &    1.662 &    0.774 \\
				& Mean ALL & -118.337 & -118.337 & -118.339 \\
				& MSE weights &        0 &        0 &    0.008 \\
				& MSE means &    0.065 &    0.071 &    0.065 \\
				& MSE cov &    0.228 &    0.235 &    0.228 \\
				\bottomrule
		\end{tabular}}
		\label{tab:e=10,d40,K5,N10000}
	\end{subtable}
	\label{tab:e=10,d40,K5}
\end{table}

%% file: Chapters/Appendix/Realworld.tex
\FloatBarrier
\subsubsection{Gas Turbine CO and NOx Emission Data Set \citep{Dataset_emission}}
\begin{table}[h]
	\footnotesize
	\centering
	\caption{Results of (normalized) Magic Gamma telescope data set for different number of components $K$. Number of observations $m=36733$, dimensions $d=11$.}
	\begin{tabular}{llrrr}
		\toprule
		&     &      EM &  R-NTR &  R-LBFGS \\
		\midrule
		K = 2 & Time (s) &    2.04 &   2.48 &     3.22 \\
		& Iterations &   25 &  22 &    28 \\
		& ALL &   -5.73 &  -5.73 &    -5.73 \\
		K = 5 & Time (s) &   17.41 &   9.52 &    11.48 \\
		& Iterations &   90 &  46 &    58 \\
		& ALL &   -1.93 &  -1.99 &    -1.93 \\
		K = 10 & Time (s) &   40.66 &  29.36 &    27.31 \\
		& Iterations &  130 &  61 &    75 \\
		& ALL &   -1.17 &  -1.16 &    -1.17 \\
		K = 15 & Time (s) &   52.06 &  63.20 &   142.77 \\
		& Iterations &  115 &  76 &   233 \\
		& ALL &    0.04 &  -0.04 &    -0.05 \\
		\bottomrule
	\end{tabular}
	\label{tab:Emission dataset}
\end{table}

\subsubsection{Wine Quality Data set \citep{Dataset_winequal}}
\begin{table}[h]		
\footnotesize
\centering
	\caption{Results of (normalized) Wine quality data set for different number of components $K$. Number of observations
			$m=6497$
		, dimensions $d=11$.}
	\begin{tabular}{llrrr}
		\toprule
		&     &       EM &  R-NTR &  R-LBFGS \\
		\midrule
		K = 2 & Time (s) &     0.54 &   0.24 &     1.55 \\
		& Iterations &    27 &   8 &    20 \\
		& ALL &   -11.02 & -11.02 &   -11.02 \\
		K = 5 & Time (s) &     4.75 &   1.24 &     5.69 \\
		& Iterations &   154 &  34 &    51 \\
		& ALL &    -9.74 &  -9.98 &    -9.88 \\
		K = 10 & Time (s) &    12.61 &   5.65 &    22.80 \\
		& Iterations &   239 &  83 &   100 \\
		& ALL &    -9.23 &  -9.28 &    -9.28 \\
		K = 15 & Time (s) &    91.16 &  11.17 &    51.65 \\
		& Iterations &  1137 &  70 &   147 \\
		& ALL &    -8.91 &  -8.88 &    -8.89 \\
		\bottomrule
	\end{tabular}	
	\label{tab:WineQuality}
\end{table}

The wine quality data set also provides classification labels: we can distinguish between white and red wine or distinguish between 7 quality labels. Besides the clustering performance of the methods (Table \ref{tab:WineQuality}), we also show the goodness of fit of our method for $K=2$ and $K=7$.

\begin{table}[h]	
	\caption{Model quality for (normalized) wine data set for $K=2$ (red/white wine).}
	\begin{subtable}[t]{.49\textwidth}
		\caption{Weighted mean squared errors of\\ (normalized) wine data set for $K=2$.\\ Weighting by mixing coefficients\\ of the respective method.}
		\resizebox{!}{.15\textwidth}{
		\begin{tabular}{lrrr}
			\toprule
			{} &        EM &     R-NTR &   R-LBFGS \\
			\midrule
			distance    &  2.757966 &  2.757982 &  2.757974 \\
			wMSE weight &  0.003169 &  0.003169 &  0.003169 \\
			wMSE mean   &  0.073776 &  0.073779 &  0.073778 \\
			wMSE cov    &  0.562106 &  0.562113 &  0.562109 \\
			\bottomrule
	\end{tabular}}
	\label{tab:WineQuality_gof2}
	\end{subtable}
	\begin{subtable}[t]{.49\textwidth}
		\caption{Adjusted Rand Index for (normalized) wine data set for $K=2$. \\[3ex] }
		\resizebox{!}{.15\textwidth}{
			\begin{tabular}{llllr}
				\toprule
				{} & truth &    EM & R-NTR &  R-LBFGS \\
				\midrule
				truth &            1 &  0.02 &  0.01 &     0.02 \\
				EM           &              &     1 &  0.73 &     0.70 \\
				R-NTR        &              &       &     1 &     0.85 \\
				R-LBFGS      &              &       &       &     1 \\
				\bottomrule
		\end{tabular}}
	\label{tab:ari_wine_2}
	\end{subtable}
\end{table}

\begin{table}[h]	
	\caption{Model quality for (normalized) wine data set for $K=7$ (quality label).}
	\begin{subtable}[t]{.49\textwidth}
		\centering
		\caption{Weighted mean squared errors of\\ (normalized) wine data set for $K=7$.\\ Weighting by mixing coefficients\\ of the respective method.}
		\resizebox{!}{.15\textwidth}{
		\begin{tabular}{lrrr}
			\toprule
			{} &         EM &      R-NTR &    R-LBFGS \\
			\midrule
			distance    &  89.259333 &  89.488919 &  88.894978 \\
			wMSE weight &   0.019019 &   0.010065 &   0.035496 \\
			wMSE mean   &   6.068149 &   5.617490 &  11.996024 \\
			wMSE cov    &  61.808676 &  37.060684 &  55.707085 \\
			\bottomrule
		\end{tabular}}
		\label{tab:WineQuality_gof7_MethodenGew}
\end{subtable}
\begin{subtable}[t]{.49\textwidth}
	\footnotesize
	\centering
	\caption{Adjusted Rand Index for (normalized) wine data set for $K=7$. \\[3ex] }
	\resizebox{!}{.15\textwidth}{
		\begin{tabular}{llllr}
			\toprule
			{} & truth &    EM & R-NTR &  R-LBFGS \\
			\midrule
			truth &            1 &  0.02 &  0.01 &     0.02 \\
			EM           &              &     1 &  0.73 &     0.70 \\
			R-NTR        &              &       &     1 &     0.85 \\
			R-LBFGS      &              &       &       &     1 \\
			\bottomrule
		\end{tabular}}
\end{subtable}
\end{table}
